\documentclass{article}

\usepackage[final, main, nonatbib]{neurips_2026}

\usepackage[numbers,sort&compress]{natbib}
\usepackage[utf8]{inputenc}
\usepackage[T1]{fontenc}
\usepackage{hyperref}
\usepackage{url}
\usepackage{booktabs}
\usepackage{amsfonts}
\usepackage{amsmath}
\usepackage{nicefrac}
\usepackage{microtype}
\usepackage{xcolor}
\usepackage{graphicx}
\usepackage{subcaption}
\usepackage{xspace}
\usepackage{listings}
\usepackage{arydshln}
\usepackage{wrapfig}
\usepackage{titlesec}
\usepackage{algorithm}
\usepackage{algpseudocode}
\usepackage{pgfplots}
\pgfplotsset{compat=1.17}
\usepgfplotslibrary{groupplots}

\definecolor{oiblue}{RGB}{0,114,178}
\definecolor{oiorange}{RGB}{230,159,0}
\definecolor{oigreen}{RGB}{0,158,115}
\definecolor{oired}{RGB}{213,94,0}
\definecolor{oipurple}{RGB}{204,121,167}
\definecolor{oisky}{RGB}{86,180,233}

\newcommand{\sysname}{Primus Projection\xspace}
\newcommand{\agentname}{Primus Tuning Agent\xspace}

\titlespacing*{\section}{0pt}{1.0ex plus 0.4ex minus 0.2ex}{0.5ex plus 0.2ex}
\titlespacing*{\subsection}{0pt}{0.7ex plus 0.3ex minus 0.2ex}{0.3ex plus 0.2ex}
\titlespacing*{\paragraph}{0pt}{0.4ex plus 0.2ex}{0.6em}
\title{Beyond the Trial-and-Error Loop: Hybrid Projection and Automated Tuning for Distributed Training}

\author{%
  {\bf Anshu Raina \quad Peyman Razaghi \quad Yuankai Chen \quad Cheng Yao \quad Yao Fu}\\
  {\bf Vidushi Goyal \quad Devang Patel \quad Wen Xie \quad Zhenyu Gu \quad Emad Barsoum}\\[4pt]
  Advanced Micro Devices, Inc.\\[2pt]
  Correspondence: \texttt{anshu.raina@amd.com}
}

\hypersetup{
  pdftitle={Beyond the Trial-and-Error Loop: Hybrid Projection and Automated Tuning for Distributed Training},
  pdfauthor={Anshu Raina, Peyman Razaghi, Yuankai Chen, Cheng Yao, Yao Fu, Vidushi Goyal, Devang Patel, Wen Xie, Zhenyu Gu, Emad Barsoum},
}

\begin{document}

\maketitle

\begin{abstract}
Configuring distributed large language model (LLM) training across
hundreds of graphics processing units (GPUs) requires
jointly optimizing tensor, pipeline, expert, context, and data
parallelism alongside micro-batch size, recomputation, sharding, and
overlap, a combinatorial space where the boundary between
memory-legal and out-of-memory (OOM) can hinge on a single byte of per-element
activation accounting. Today, teams navigate it through expensive
multi-node trial and error, burning GPU-hours on crashes and
suboptimal recipes. We present \emph{\sysname}, which closes this loop with
two coupled components. First, a hybrid projection engine that
measures what it can and simulates what it cannot: GPU microbenchmarks
on as few as a single device capture compute kernels and any in-scope
communication; a sub-node downscale--measure--upscale workflow
analytically restores out-of-scope collectives, fully sharded data
parallel (FSDP) overlap, and the full pipeline schedule at the
multi-node target. A pure simulation mode runs entirely on a central
processing unit (CPU), enabling pre-silicon planning without any
GPU. Second, the \emph{\agentname} treats the projection engine as a fast
scoring oracle (seconds per candidate instead of tens of
minutes) and uses it to drive an intelligent search over the
configuration space. A deterministic seed planner sweeps single axes;
an LLM investigation loop proposes cross-axis combinations the planner
cannot reach, with an analytic memory pre-filter that rejects
infeasible candidates before any GPU time is spent. Validated on
dense Llama~3.1 and mixture-of-experts (MoE) Mixtral~8$\times$22B
across two GPU generations, all projections stay within $\sim$10\% of measured
multi-node throughput, and fidelity holds at production scale: 4.7\%
error on DeepSeek~V3 671B at 1{,}024~GPUs and 1.9\% on
Qwen3-235B-A22B at 256~GPUs. In a Mixtral~8$\times$22B case study, the
\emph{\agentname} discovers a configuration delivering +27\% throughput
over the published 4-node bfloat16 (BF16) reference at the same scale, in under
30~minutes of single-node exploration, with no hand-written configs
and no full-cluster profiling pass.
\end{abstract}

\section{Introduction}
\label{sec:intro}

The rapid growth in large language model (LLM) scale, from billions
to trillions of parameters, has made distributed training
configuration a first-class engineering challenge. A single 100B+ run
jointly selects five interacting parallelism dimensions---tensor
parallelism (TP), pipeline parallelism (PP), expert parallelism (EP),
context parallelism (CP), and data parallelism (DP)---plus micro-batch
size, sequence length, gradient accumulation, recomputation, FSDP
sharding, and overlap flags; each combination
trades memory, compute utilization, and communication overhead
differently, and the combinatorial space of valid configurations grows
rapidly with model size and cluster scale, making exhaustive
trial-and-error infeasible. In practice, teams launch full multi-node
runs, observe OOMs or suboptimal throughput, adjust, and repeat---each
iteration consuming hours of expensive cluster time and delaying model
development.

Existing tools fall short of closing this loop. Analytical cost models
(e.g., Calculon~\cite{calculon}, Galvatron~\cite{galvatron}) predict
training time from hardware specifications or single-device layer
profiles with nominal-bandwidth communication, but miss GPU frequency
throttling under sustained load, software-stack variation, and
grouped general matrix multiplication (GEMM) behavior in
mixture-of-experts (MoE) models~\cite{gshard,switchtransformer,megablocks}. Full-system
simulators (e.g., ASTRA-sim~\cite{astrasim}, Echo/Moye~\cite{echo}) achieve
higher fidelity but require target-hardware traces or extensive
calibration, and do not scale to the configuration sweeps needed for
planning.

We present \textbf{\sysname}: compute kernels are benchmarked on as few as a
single GPU; out-of-scope communication, FSDP overlap, and the full
pipeline schedule are reconstructed analytically at the target scale.
A pure simulation mode runs entirely on CPU for pre-silicon planning.
The \textbf{\agentname} treats the projection oracle as a fast surrogate to
search the full configuration space end-to-end---parallelism,
schedule, recomputation, MoE backend, and precision---replacing days
of manual profiling with minutes of single-node exploration.

We validate \sysname against published multi-node training
results~\cite{amd-perf} on AMD Instinct\texttrademark{} MI325X and MI355X GPUs across
dense (Llama~3.1~\cite{llama3}) and MoE (Mixtral~8$\times$22B~\cite{mixtral})
models, keeping all projections within $\sim$10\% of measured
throughput, and at production scale on DeepSeek~V3
671B~\cite{deepseekv3} (1{,}024~GPUs, 4.7\% error) and
Qwen3-235B-A22B~\cite{qwen3} (256~GPUs, 1.9\%). Projection also
preserves configuration rankings (Spearman $\rho\ge0.93$) on Mixtral,
Qwen3-235B-A22B, and GPT-OSS-120B~\cite{gptoss}, always selecting the
measured-best configuration. In a Mixtral tuning case
study, the agent delivers +27\% throughput over the published AMD 4-node
BF16 reference at the same scale in under 30~minutes.

\section{Background and Motivation}
\label{sec:background}

A distributed LLM training run is defined by a configuration vector
\textbf{(TP, PP, VPP, EP, CP, DP)}, where virtual pipeline parallelism
(VPP) is the interleaving degree, along with choices for micro-batch
size (MBS), global batch size (GBS), sequence length, gradient
accumulation, recomputation, FSDP
sharding, and communication overlap. Each combination determines a
unique point in a high-dimensional space of memory consumption,
compute utilization, communication overhead, and pipeline bubble
fraction. For a Mixtral~8$\times$22B run on 64~GPUs, the partition
constraint $\text{TP}\times\text{PP}\times\text{EP}\times\text{CP}
\le 64$ alone admits roughly 160 ordered factorizations; layered with
$\text{MBS}\in\{1,\ldots,8\}$, recomputation level
$\in\{\text{none, sel, full}\}$, MoE backend $\in\{$All-to-All (A2A),
DeepEP$\}$, and pipeline algorithm $\in\{$one-forward-one-backward
(1F1B), Interleaved~\cite{megatron-v2}, zero bubble (ZB)~\cite{zerobubble},
zero bubble V-schedule (ZBV), integer linear programming (ILP)$\}$,
the search space grows to $\mathcal{O}(10^4)$ topologically valid
points, of which only a small fraction is simultaneously memory-legal
and high-throughput---and the boundary between feasible and infeasible
is razor-thin: a 4-byte miscount of activation memory per element
across 56 MoE layers is enough to push a configuration from
\emph{legal} to \emph{OOM}, voiding hours of queued cluster time.
Manual exploration in this space is impractical.

Pure analytical performance modeling has inherent pitfalls: GPU clock throttling under load creates a 44--70\% gap between peak and maximum achievable floating-point operations per second (FLOPS)~\cite{amd-maf}; AMD ROCm\texttrademark{} software updates (hipBLASLt, Composable Kernel, Triton) shift per-operation throughput unpredictably; MoE grouped GEMMs depend on wave quantization and routing distributions that resist closed-form treatment~\cite{megablocks}; and optimizer, allocator, and gradient-accumulation overhead is invisible to roofline models. Benchmarking actual kernels on the deployed stack captures all four automatically. \sysname pairs such a benchmark---measured on as few as a single
GPU---with analytical models for collective communication, pipeline
scheduling, and FSDP overlap, composing the two so that the projection
transfers to any multi-node target without launching a multi-node run.

\section{Methodology}
\label{sec:method}

\begin{figure}[h]
\centering
\includegraphics[width=0.95\linewidth]{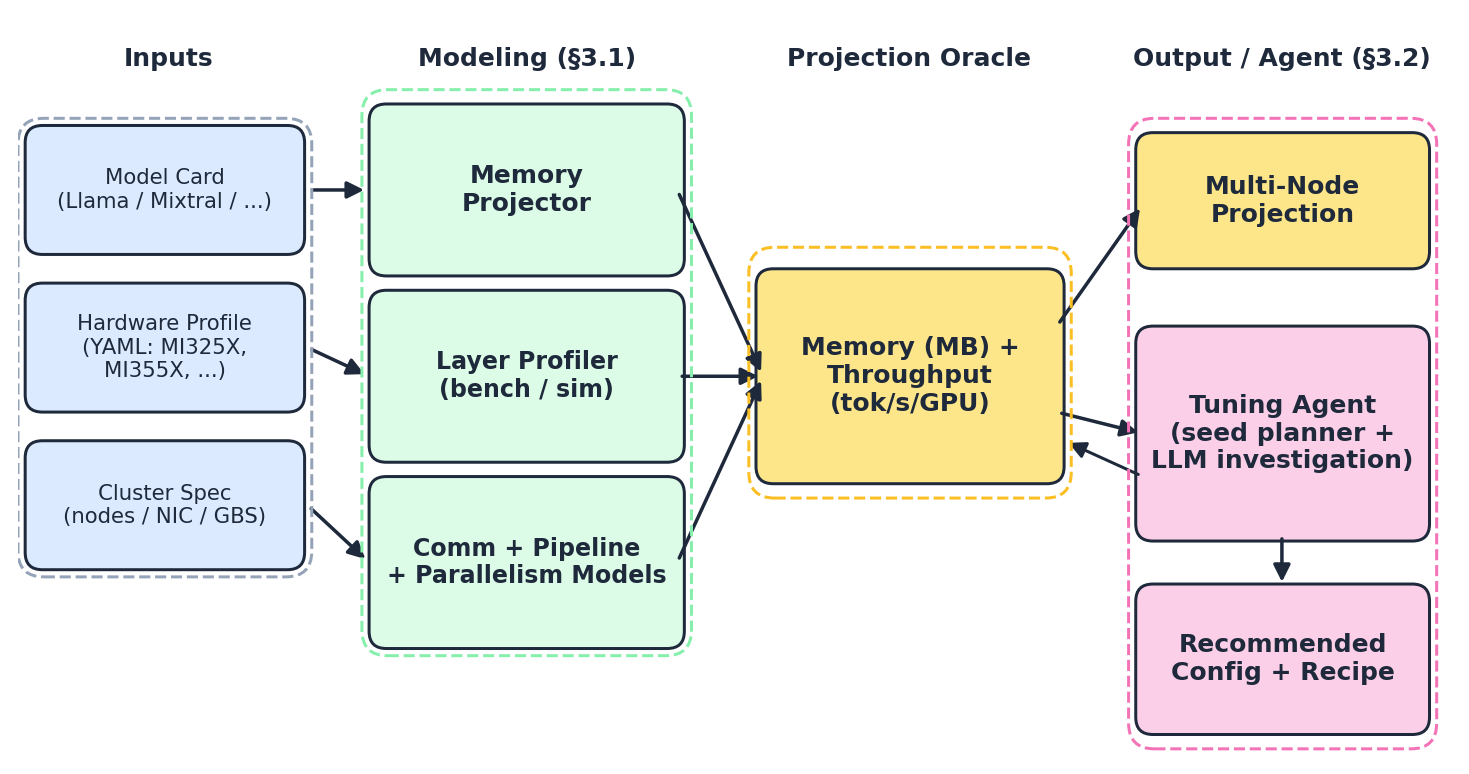}
\caption{\small \sysname architecture. Modeling components consume model card,
hardware profile, and cluster spec; the \emph{projection oracle}
returns a memory and throughput estimate that drives both multi-node
projection and the \agentname's iterative search.}
\label{fig:method-overview}
\end{figure}

\sysname is a modular pipeline of five components: memory projection
(analytical), layer profiling (benchmark or simulation), communication
modeling, pipeline schedule simulation, and multi-node scaling.
Hardware profiles for AMD Instinct MI300X, MI325X, and MI355X GPUs are
bundled and custom YAML profiles can be added (Appendix~\ref{app:env}).
The two top-level capabilities---\emph{Modeling} and the
\agentname built on it---form a single projection oracle
(Figure~\ref{fig:method-overview}).

\subsection{Modeling}
\label{sec:method-model}

\paragraph{Memory:} A hierarchical profiler tree mirroring the model's
module structure (embeddings, dense / MoE transformer layers, output,
loss) returns sharded per-GPU parameter, gradient, and optimizer-state
counts together with per-microbatch activation memory, accounting for
TP weight splits, PP layer assignment, EP expert sharding, CP
replication, and FSDP/Zero Redundancy Optimizer (ZeRO) sharding across
DP. The projector treats precision as two independent axes---a
\emph{parameter/optimizer} regime (BF16 vs.\ 8-bit floating-point (FP8)
mixed precision; Adam / AdamW / distributed optimizer)
and a \emph{saved-activation} regime (BF16 by default, FP8 when the
framework casts saved activations to FP8 for memory savings)---each
contributing a per-element byte cost. Pipeline-schedule interleaving,
gradient accumulation, and recomputation (full or selective) compose
with the activation regime; full recomputation reduces the per-layer
activation footprint to a single checkpointed input, again at the
selected activation precision. Activation memory dominates for large
MoE workloads (a single MoE layer can outweigh attention activations
by an order of magnitude), so per-layer accounting must be
accurate. Per-component
equations and byte costs are in Appendix~\ref{app:models}.

\paragraph{Performance: measure what you can, simulate what you can't.}
\sysname's default mode combines both: when a parallelism dimension
fits inside the available benchmark GPUs---TP AllReduce within a
node, EP All-to-All within the benchmark configuration, intra-node DP
gradient AllReduce---the corresponding collective is captured
directly by the kernel-level benchmark alongside compute. Only
communication that falls outside the single-node benchmark scope,
primarily inter-node traffic and the deltas introduced when a
parallelism dimension is reduced for benchmarking, is reconstructed
analytically. In the limit of a single-GPU benchmark no collective is
directly measurable, since collectives are inherently multi-GPU; in
that case every communication term is reconstructed from the analytic
model below, and only compute time is measured.
Pipeline scheduling follows deterministic dependency rules and is
simulated exactly from the per-stage compute times produced by the benchmark.
Framework overhead (allocator pauses, gradient-accumulation glue) is
absorbed implicitly into the measured kernel sequence; the optimizer
step runs once per iteration and is measured directly on the benchmark
(falling back to an analytic high-bandwidth memory (HBM) bandwidth model). Per-iteration time
composes as
\begin{equation}
\text{iter\_time}=
T_{\text{pipe}}\big(\{t_\ell\},\,N_{\mu\text{B}},\,\text{schedule}\big)
+\text{FSDP}^{\text{exp}}
+\text{grad\_AR}^{\text{exp}}
+\text{opt\_step},
\label{eq:itertime}
\end{equation}
where $t_\ell$ is the per-layer forward/backward time (measured, plus the
analytic TP/EP deltas of Algorithm~\ref{alg:subnode}; in-scope TP
AllReduce and EP All-to-All are therefore already inside $t_\ell$),
$T_{\text{pipe}}$ is the simulated makespan of the target pipeline schedule
over $N_{\mu\text{B}}$ micro-batches (including bubble and inter-stage point-to-point (P2P) transfers;
it reduces to $N_{\mu\text{B}}\sum_\ell t_\ell$ when $\text{PP}{=}1$), and the
superscript ``exp'' denotes only the exposed, non-overlapped portion.

\paragraph{Sub-node benchmarking: downscale, measure, upscale.}
\label{sec:subnode}
The \texttt{-{}-benchmark-gpus} flag fixes the benchmark hardware budget
(defaulting to one node); the projector then follows a
\emph{downscale $\to$ measure $\to$ upscale} workflow
(Algorithm~\ref{alg:subnode}).

\begin{algorithm}[h]
\caption{\small Sub-node benchmarking: downscale $\to$ measure $\to$ upscale.}
\label{alg:subnode}
\small
\begin{algorithmic}[1]
\Require target config $C^{*}=(\text{TP},\text{PP},\text{VPP},\text{EP},\text{CP},\text{DP})$;
benchmark budget $G_b$; hardware profile $H$
\State $C \gets C^{*}$
\Statex \textbf{Phase 1: Downscale} \Comment{shrink along axes by restoration-fidelity order}
\While{$\textsc{GpuCount}(C) > G_b$}
  \If{$C.\text{PP}>1$} $C.\text{PP} \gets 1$ \Comment{schedule replayed exactly at upscale}
  \ElsIf{$C.\text{EP}>1$} shrink $C.\text{EP}$, scale \texttt{num\_experts} so experts/rank is preserved
  \Else{} shrink $C.\text{TP}$ \Comment{slice shape changes; compute is rescaled at upscale}
  \EndIf
  \Statex \quad ($C.\text{CP}$ is never reduced)
\EndWhile
\State $C^{\text{bench}} \gets C$ \Comment{downscaled config actually run on the $G_b$ GPUs}
\Statex \textbf{Phase 2: Measure} \Comment{minimal layer set; one timing per layer type}
\For{each distinct layer type $\ell\in\{\text{dense},\text{moe}\}$ in the model}
  \State measure forward and backward times $(t_\ell^{F}, t_\ell^{B})$ on $G_b$ GPUs
  \If{$\ell$ is MoE} additionally split $t_\ell$ into compute and All-to-All components
  \EndIf
\EndFor
\Statex \textbf{Phase 3: Upscale} \Comment{compose analytical deltas onto measured kernels}
\State PP: simulate target schedule from $\{t_\ell\}$, add intra-/inter-node P2P
\State EP: per MoE layer, add $\Delta_{\text{A2A}}=\text{A2A}(C^{*})-\text{A2A}(C^{\text{bench}})$
\State TP: rescale compute by $\text{TP}^{\text{bench}}/\text{TP}^{*}$, add AllReduce (AR) delta
\Statex \quad (with measured AR, use ratio form $\text{AR}^{*}=\text{AR}^{\text{bench}}\times a^{*}/a^{\text{bench}}$)
\State \Return $\text{iter\_time}$ via Eq.~\ref{eq:itertime}; tok/s/GPU
\end{algorithmic}
\end{algorithm}

\emph{Downscale.} When the target requires more GPUs than the
benchmark host provides, parallelism dimensions are reduced in a
priority order $\text{PP}\to\text{EP}\to\text{TP}$ chosen to minimize
analytical restoration error. \textbf{PP$\to$1} comes first because
pipeline parallelism's effect on iteration time is entirely
determined by the schedule, which is deterministically simulable;
removing PP for benchmarking loses no compute information.
\textbf{EP is reduced next}, with \texttt{num\_experts} adjusted
proportionally so that each GPU still computes the same number of
experts---the per-rank MoE workload (and therefore the measured GEMM
time) is preserved, leaving only the All-to-All to be reconstructed.
\textbf{TP is reduced last} because shrinking it changes the per-GPU
tensor slice shape, forcing both a compute-time rescale and an
AllReduce delta. \textbf{CP is never reduced}: in MoE workloads CP
is folded into EP (MoE Parallel Folding), so shrinking it saves no
GPUs; in dense workloads it would change the per-GPU sequence length
and fundamentally alter GEMM shapes and attention computation, so the
benchmark would no longer measure the target workload. In practice
this is rarely a constraint, since most dense models train at 8K--32K
sequence lengths where CP=1. As a concrete example,
a target with $(\text{TP}{=}1,\text{PP}{=}4,\text{EP}{=}8,
\text{CP}{=}1)$ requiring 32 GPUs downscales to a 1-GPU benchmark in
two steps: PP $4{\to}1$ (8~GPUs), then EP $8{\to}1$ (1~GPU).

\emph{Measure.} The projector benchmarks a minimal set of layers
covering every distinct layer \emph{type}: one dense layer for a
pure dense model, one MoE layer for an all-MoE model, or one of each
for mixed architectures (e.g., DeepSeek's 3~dense $+$ 58~MoE layers).
All layers of the same type share architecture and GEMM shapes, so
differences in learned weights do not affect kernel time, and each
selected layer is timed for forward and backward separately. For MoE
layers a \emph{decomposed measurement} captures compute and All-to-All
times independently, because A2A will be replaced analytically during
upscale while the compute portion is kept verbatim. When PP$=$1 the
per-type time scaled by the layer count of that type yields the
full-model time; when PP$>$1 the pipeline simulator constructs a
per-stage time matrix from the model's layer assignment (respecting
\texttt{decoder\_first}/\texttt{last\_pipeline\_num\_layers} and
\texttt{pipeline\_model\_parallel\_layout}) and simulates the schedule
with exact bubble accounting.

\emph{Upscale.} Each reduced dimension is restored analytically and
the contributions compose. PP is reconstructed by simulating the
target schedule (1F1B, Interleaved, Zero-Bubble, ZBV, ILP) from
benchmarked per-layer times, distinguishing intra- and inter-node
P2P. EP is restored by adding, per MoE layer, the A2A delta
$\Delta=\text{A2A}(C^{*})-\text{A2A}(C^{\text{bench}})$. Both terms
come from the topology-aware collective model
(Appendix~\ref{app:models}): $\text{A2A}(C^{\text{bench}})$ is the A2A
of the downscaled configuration that was actually benchmarked, and
$\text{A2A}(C^{*})$ is the A2A of the full target EP group, priced on
inter-node links whenever that group spans nodes. When the decomposed
MoE measurement is available, the measured intra-node A2A anchors the
model, so only the change from the benchmark group to the target group
is extrapolated. TP applies two adjustments per
layer: compute is rescaled by $\text{bench\_tp}/\text{target\_tp}$
and a TP AllReduce delta is added; with measured AllReduce data, the
ratio form
$\text{target}_{\text{AR}}{=}\text{measured}_{\text{AR}}\times
[\text{analytical}(\text{target})/\text{analytical}(\text{bench})]$
improves accuracy. The order $\text{PP}\to\text{EP}\to\text{TP}$ is
ranked by restoration fidelity: PP is exact (only the P2P latency
model contributes residual), EP preserves compute (only A2A is
exercised), and TP relies on a compute-scaling assumption that holds
well for dense GEMMs but is the least exact of the three.

\emph{Profiling modes.} Independently of the downscale\,/\,measure\,/\,upscale
loop, the projector exposes a \texttt{simulate} mode that runs without
any GPU---layer kernel times come from the Origami GEMM
model~\cite{origami} and a tile-level Flash~Attention~v3
simulator~\cite{flashattn3}---and a \texttt{both} mode that runs
benchmark and simulation side-by-side for cross-validation. The
simulate mode supports laptop-scale planning and pre-silicon
exploration of unreleased GPU profiles.

\paragraph{Communication, pipeline, and parallelism:}
The communication model spans three topology domains---intra-node,
inter-node, inter-pod---each parameterized by per-link bandwidth and
latency drawn from the hardware-profile YAML, so the same engine
projects MI300X inter-chip global memory interconnect (xGMI) clusters, MI355X scale-up pods, or hypothetical
future fabrics without code changes. For each collective the model
enumerates the applicable algorithms (AllReduce: bandwidth-optimal
ring~\cite{ringallreduce}, hypercube, Bruck, single-shot $\times$ four
protocols; All-to-All: direct, hierarchical, PXN/DeepEP; P2P:
latency+bandwidth) and selects the minimum-time candidate for the
target topology and message size. For
MoE models with EP$>$1, gradient AllReduce is decomposed into expert
(across DP/EP replicas) and non-expert components, with expert AR on
the critical path because A2A barriers prevent overlap with backward
compute. FSDP communication uses a per-layer overlap model with
phase-specific ceilings. The pipeline simulator supports six scheduling
algorithms (1F1B, Interleaved 1F1B, Zero-Bubble,
ZBV Formatted/Greedy, Sea AI Lab ILP~\cite{zerobubble}); a
\emph{compare-all} flag runs every applicable scheduler. Reference
bandwidth/latency values, scheduler bubble formulas, and FSDP
ceilings are in Appendix~\ref{app:models}.

\subsection{\agentname}
\label{sec:method-agent}

The configuration space is jointly optimized over parallelism
dimensions, pipeline schedule, micro-batch size, recomputation
granularity, precision mode, MoE communication backend, and overlap
flags. The agent treats the projection oracle as a cheap analytical
surrogate (seconds per candidate on a single node) and converts the
expensive multi-node search into a programmatic one. Given a model
card, target cluster, and HBM budget, it returns a high-throughput,
memory-legal recipe end-to-end with no hand-written Megatron configs.

The agent operates in two phases over a shared trial ledger that serves
as its session memory. A
deterministic \emph{seed planner} emits 12 candidates by walking
single-axis sweeps (memory frontier; MoE backend; pipeline schedule
and VPP depth; micro-batch size and distributed optimizer); each axis
is anchored on the user-supplied baseline so that single-axis
improvements are directly attributable. The seed phase establishes a
ledger of \emph{(configuration, projected throughput, projected
memory, legality)} tuples that grounds the second phase. An \emph{LLM
investigation} loop then reads this ledger and proposes cross-axis
combinations as structured JSON, evaluated iteratively by the same
projection oracle (up to 25 trials per round across up to 4 rounds). The
loop, structured in the spirit of ReAct~\cite{react}-style interleaved
reasoning and recursive tool use~\cite{rlm}, is implemented in
DSPy~\cite{dspy} as two cooperating modules: a \emph{Planner LLM} emits
a hypothesis paragraph plus a JSON list of candidate configurations to
evaluate next, and a \emph{Recursive Language Model}
(RLM)~\cite{rlm} executes those candidates one at a time, calling
the projection oracle as a tool, writing notes to a per-round
scratchpad, and optionally querying a sub-LLM for deeper analysis of
intermediate results. Every candidate first passes an analytic memory
pre-filter that rejects OOM-bound configurations before any GPU time
is consumed ($\sim$25~minutes saved per rejection); survivors are
scored via sub-node benchmark projection. A trial is \emph{legal} iff
axis validation passes ($\text{PP}\!\times\!\text{DP}\!\times\!
\text{CP}\!\times\!\text{TP}$ tiles the cluster), projected memory
$\le$ cap, and the benchmark returns a parseable throughput; the
winner is $\arg\max_{\text{legal}}(\text{tok/s/GPU})$. The seed
schedule, LLM prompting strategy, and trial-scoring rule are
documented in Appendix~\ref{app:agent}.

\section{Evaluation}
\label{sec:eval}

\subsection{Methodology and Sub-Node Sweep}

We validate \sysname against measured throughput on the AMD ROCm
software performance results page~\cite{amd-perf}. The workflow benchmarks with
\textbf{1, 2, 4 GPUs and 1 node (=\,8 GPUs)} on the same host, then
projects to a multi-node target using a hardware-profile YAML; using
fewer GPUs for the benchmark shifts more of the communication stack
into the analytical model. Dense models (Llama~3.1~\cite{llama3})
target 8~nodes (64~GPUs on MI325X); the MoE model
(Mixtral~8$\times$22B~\cite{mixtral}) targets 4~nodes (32~GPUs on
MI355X). Tables~\ref{tab:dense} and~\ref{tab:moe} summarize the sweep;
Figure~\ref{fig:heatmap} visualizes all 16 cells.

\begin{table}[h]
\centering
\scriptsize
\setlength{\tabcolsep}{2pt}
\begin{tabular}[c]{ll r r r l r r@{\,}l r@{\,}l r@{\,}l r@{\,}l c}
\toprule
\textbf{Model} & \textbf{Prec.} & \textbf{MBS} & \textbf{GBS} & \textbf{Seq} & \textbf{FSDP}
  & \textbf{Meas.}
  & \multicolumn{2}{c}{\textbf{1\,GPU}}
  & \multicolumn{2}{c}{\textbf{2\,GPUs}}
  & \multicolumn{2}{c}{\textbf{4\,GPUs}}
  & \multicolumn{2}{c}{\textbf{8\,GPUs}}
  & \textbf{Max\,$\Delta$} \\
\midrule
Llama 8B  & FP8  & 2 & 128 & 8192 & No
  & 16,186
  & 17,619 & \scriptsize{(+8.9\%)}
  & 17,628 & \scriptsize{(+8.9\%)}
  & 17,604 & \scriptsize{(+8.8\%)}
  & 17,607 & \scriptsize{(+8.8\%)}
  & 0.1\% \\
Llama 70B & FP8  & 4 & 256 & 8192 & Yes
  & 1,726
  & 1,867 & \scriptsize{(+8.2\%)}
  & 1,868 & \scriptsize{(+8.2\%)}
  & 1,850 & \scriptsize{(+7.2\%)}
  & 1,825 & \scriptsize{(+5.7\%)}
  & 2.4\% \\
Llama 70B & BF16 & 1 & 64  & 8192 & Yes
  & 1,174
  & 1,087 & \scriptsize{($-$7.4\%)}
  & 1,091 & \scriptsize{($-$7.1\%)}
  & 1,087 & \scriptsize{($-$7.4\%)}
  & 1,086 & \scriptsize{($-$7.5\%)}
  & 0.5\% \\
\bottomrule
\end{tabular}
\caption{\small Dense model sub-node benchmarking sweep (MI325X, target: 8~nodes / 64~GPUs, tok/s/GPU). Parenthetical percentages are error vs.\ measured. Max~$\Delta$ = max deviation across benchmark GPU counts relative to the 8-GPU projection.}
\label{tab:dense}
\end{table}

\begin{figure}[t]
\centering
\begin{minipage}[t]{0.60\linewidth}
  \captionof{figure}{\small Projection error heatmap---all 16 cells within $\sim$10\% of measured. Color encodes $|$error$|$ (Okabe-Ito; green $=$ near zero); labels are signed ($+$ $=$ over-projection).}
  \label{fig:heatmap}
  \vspace{0.3em}
  \includegraphics[width=\linewidth]{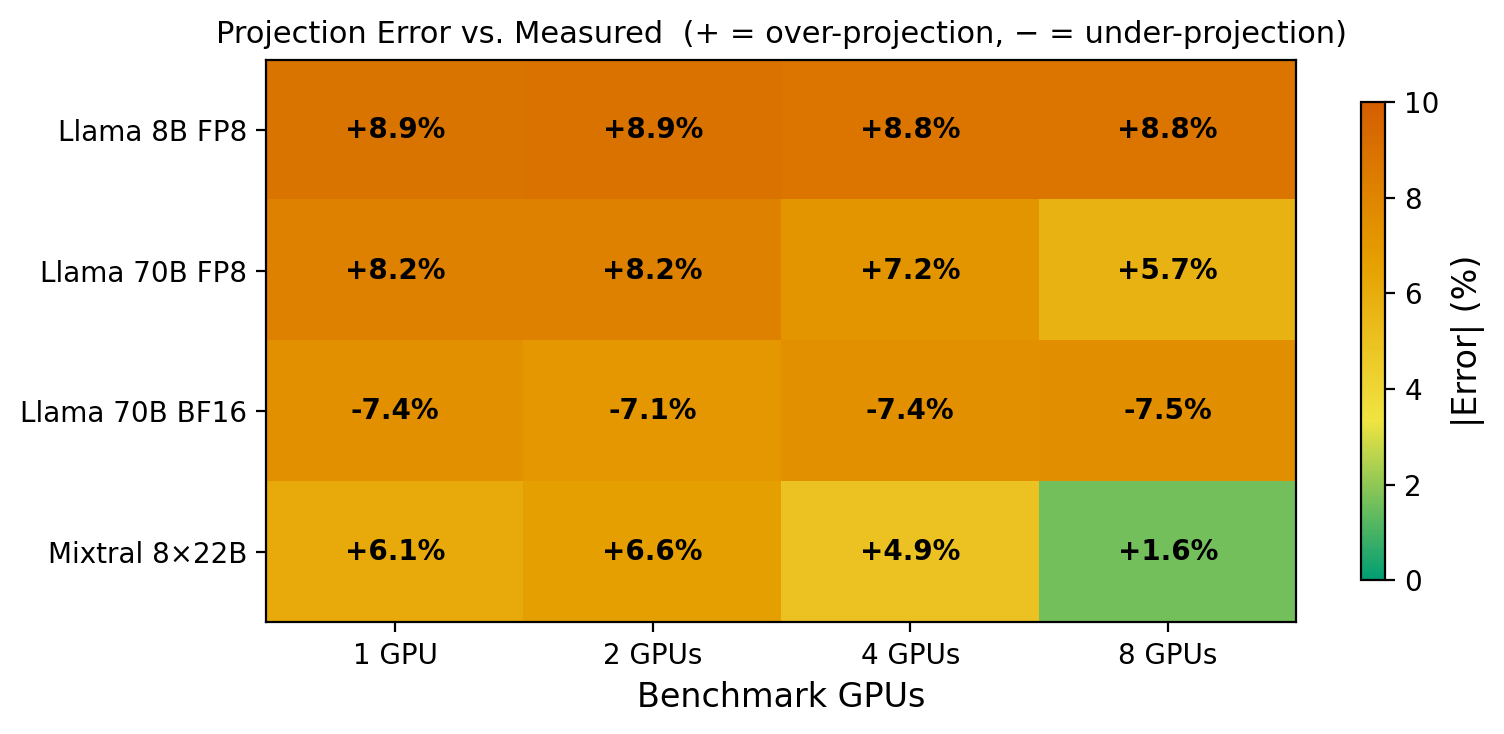}
\end{minipage}%
\hspace{0.03\linewidth}%
\begin{minipage}[t]{0.35\linewidth}
  \captionof{table}{\small Communication model accuracy vs.\ preflight (MI325X, $\ge$64\,MB).}
  \label{tab:comm-corr}
  \vspace{0.4em}
  \scriptsize\setlength{\tabcolsep}{2pt}%
  \begin{tabular}{lc}
  \toprule
  \textbf{Collective} & \textbf{Avg |Err| $\ge$64\,MB} \\
  \midrule
  Intra AR 2/4/8\,GPUs   & 4.1 / 2.0 / 5.6\% \\
  Intra A2A 2/4/8\,GPUs  & 3.2 / 3.5 / 4.7\% \\
  Inter AR 2/4\,Nodes    & 2.1 / 5.8\% \\
  Inter A2A 2/4\,Nodes   & 5.5 / 5.5\% \\
  Inter P2P 2\,Nodes     & 3.9\% \\
  \midrule
  \textbf{Overall Avg}   & \textbf{4.2\%} \\
  \bottomrule
  \end{tabular}
\end{minipage}
\end{figure}

\begin{table}[t]
\centering
\begin{minipage}[t]{0.46\linewidth}
  \captionof{table}{\small MoE sub-node sweep (Mixtral~8$\times$22B, MI355X, BF16, 4~nodes/32~GPUs, measured: 3,475 tok/s/GPU).}
  \label{tab:moe}
  \centering\scriptsize\setlength{\tabcolsep}{3pt}
  \begin{tabular}{l rrrr c}
  \toprule
  & \textbf{1\,GPU} & \textbf{2\,GPUs} & \textbf{4\,GPUs} & \textbf{8\,GPUs} & \textbf{Max\,$\Delta$} \\
  \midrule
  Mixtral & 3,688 & 3,706 & 3,647 & 3,532 & 4.9\% \\
          & {\tiny(+6.1\%)} & {\tiny(+6.6\%)} & {\tiny(+4.9\%)} & {\tiny(+1.6\%)} & \\
  \bottomrule
  \end{tabular}

  \vspace{0.5em}

  \captionof{table}{\small Per-module breakdown (rank~0). Total = emb.\ 0.617 + 1 dense (0.390) + 13 MoE (5.073) = \textbf{6.079~B}.}
  \label{tab:mem-modules}
  \centering\scriptsize\setlength{\tabcolsep}{3pt}
  \begin{tabular}{l r r}
  \toprule
  \textbf{Module (rank 0)} & \textbf{Params (B)} & \textbf{Act (GB)} \\
  \midrule
  embedding                         & 0.6166 & 0.094 \\
  dense\_transformer\_layer (1)     & 0.3901 & 1.625 \\
  \quad layer\_norm                 & $\sim$0      & 0.094 \\
  \quad self\_attention             & 0.088  & 0.313 \\
  \quad residual\_add               & 0      & 0.094 \\
  \quad mlp (dense FFN)             & 0.302  & 0.844 \\
  moe\_transformer\_layer (1)       & 0.3902 & 2.563 \\
  \quad layer\_norm                 & $\sim$0      & 0.094 \\
  \quad self\_attention             & 0.088  & 0.313 \\
  \quad residual\_add               & 0      & 0.094 \\
  \quad router                      & $\sim$0      & 0.094 \\
  \quad mlp (1 expert / rank)       & 0.302  & 1.688 \\
  final\_layernorm                  & $\sim$0      & 0.094 \\
  output\_layer (tied)              & 0.6166 & 1.531 \\
  \bottomrule
  \end{tabular}
\end{minipage}\hspace{0.04\linewidth}%
\begin{minipage}[t]{0.50\linewidth}
  \captionof{table}{\small Per-rank memory (Mixtral~8$\times$22B, MI355X, BF16, PP=4, EP=8, VPP=2, no recompute). Activation (198~GB) exceeds parameter+optimizer by $2\times$.}
  \label{tab:mem-breakdown}
  \centering\scriptsize\setlength{\tabcolsep}{4pt}
  \begin{tabular}{l r}
  \toprule
  \textbf{Bucket} & \textbf{GB} \\
  \midrule
  Param + optimizer state            &  79.26 \\
  \quad{\tiny BF16 wt + BF16 grad + FP32 Adam}         &        \\
  \quad{\tiny + master, dist.-optimizer sharded}       &        \\
  Activation (per-rank, in-flight)   & 197.83 \\
  \quad{\tiny no recompute, PP=4+VPP=2 microbatches}   &        \\
  \midrule
  \textbf{Projected total per GPU}   & \textbf{277.09} \\
  \bottomrule
  \end{tabular}

  \vspace{0.5em}

  \captionof{table}{\small Pipeline algorithm comparison on Llama2-7B (MI300X, 1~node, PP=8).}
  \label{tab:pp-llama}
  \centering\scriptsize\setlength{\tabcolsep}{2pt}
  \begin{tabular}{lcc rr r c}
  \toprule
  \textbf{Algo} & \textbf{VPP} & \textbf{tok/s} & \textbf{TFLOPs}
    & \textbf{Mem} & \textbf{Mem\%} & \textbf{Sp.} \\
  \midrule
  1F1B         & 1 & 10,057 & 235.7 & 16.26 &  8.5\% & 1.00$\times$ \\
  Interleaved  & 2 & 10,974 & 257.3 & 21.40 & 11.2\% & 1.09$\times$ \\
  ZeroBubble   & 2 & \textbf{11,411} & 268.0 & 16.59 &  8.6\% & \textbf{1.13$\times$} \\
  ZBV          & 2 & 11,347 & 265.9 & 18.11 &  9.4\% & 1.12$\times$ \\
  V-half       & 2 & 10,894 & 255.1 & 14.06 &  7.3\% & 1.08$\times$ \\
  V-min        & 2 &  8,897 & 208.5 & 11.90 &  6.2\% & 0.88$\times$ \\
  \bottomrule
  \end{tabular}
\end{minipage}
\end{table}

\subsection{Communication Model Correlation}
\label{ssec:comm-corr}

We validate the communication model in isolation against preflight
measurements on MI325X across AllReduce, All-to-All, and P2P Send/Recv
from 2~MB to 1~GB messages and intra-/inter-node topologies. Training
workloads use messages $\ge$64~MB, where the model averages
\textbf{4.2\% error} across all 11 collective/topology combinations
(Table~\ref{tab:comm-corr}); small-message latency-dominated cases
(20--30\%) are outside the training regime.

\subsection{Pipeline Schedule Algorithm Comparison}
\label{ssec:pp-algo}

Table~\ref{tab:pp-llama} compares pipeline schedules on Llama2-7B (MI300X,
1~node, PP=8). Zero-bubble achieves the highest throughput (+13\% over 1F1B)
with only 2\% memory overhead; ZBV is within 1\% at moderate memory cost;
V-half trades +8\% throughput for \emph{lower} memory than 1F1B.

\subsection{Memory Projection: Mixtral 4-Node Reference}
\label{ssec:mem-val}

Tables~\ref{tab:mem-breakdown} and~\ref{tab:mem-modules} show the per-rank
memory projection for Mixtral~8$\times$22B (4-node MI355X, BF16, PP=4, EP=8,
VPP=2, MBS=1, no recompute). Activation memory dominates at 198~GB---over
$2\times$ the parameter+optimizer budget---because EP=8 retains one expert multilayer perceptron (MLP)
per rank (0.302~B params, 1.688~GB activation) while the no-recompute setting
keeps all microbatch activations live simultaneously. This is why recompute,
MBS, and saved-activation precision are first-class axes in the agent's seed
planner, and why the memory pre-filter (which rejects 7 of 30 trials before
any GPU time is spent) is particularly valuable at MoE scale.

\subsection{End-to-End Analysis}
\label{ssec:e2e}

All projections stay within $\sim$10\% of measured throughput across
every benchmark GPU count (Figure~\ref{fig:heatmap}), spanning two
model families (dense Llama~3.1, MoE Mixtral~8$\times$22B), two
precision modes (FP8, BF16), two GPU generations (MI325X gfx942,
MI355X gfx950), and FSDP / PP=4-VPP=2 configurations
(Tables~\ref{tab:dense},~\ref{tab:moe}). The Mixtral~8$\times$22B
result exercises the full pipeline---PP reduction (PP=1$\to$4), EP A2A
modeling, VPP=2 interleaved schedule simulation, and MoE-aware gradient
AllReduce decomposition---yet the 8-GPU projection achieves only 1.6\%
error and the 1-GPU projection (where \emph{all} communication is
modeled analytically) stays within 6.1\%; on the dense side, all 12
Llama cells project the 8-node target within $\pm$9\%, with Llama~8B
FP8 the most stable (Max~$\Delta$\,=\,0.1\%) and FSDP-enabled
Llama~70B within 8.2\% (FP8) and 7.5\% (BF16) of measured. For dense
Llama the winning strategy is pure data parallelism (TP=PP=CP=1),
matching the published AMD recipe: 8B runs replicated DP, while 70B
requires full-shard FSDP2 (ZeRO-3).
The Max~$\Delta$ column quantifies restoration stability: for
configurations with simple analytical stacks (Llama~8B FP8, Llama~70B BF16)
the error is nearly independent of benchmark count (Max~$\Delta \le 0.5\%$),
leaving only a workload-level calibration bias that preserves \emph{rank ordering}
between candidates---the property the \agentname depends on. For
configurations that exercise more of the analytical stack (Llama~70B FP8
with FSDP, Mixtral PP=4 VPP=2 EP=8), error decreases as benchmark count
grows (Mixtral: $+6.1\%$ at 1~GPU to $+1.6\%$ at 8~GPUs), as direct
measurement replaces analytical components.

\subsection{Large-Scale Validation (256 and 1{,}024 GPUs)}
\label{ssec:scale}

To test whether fidelity holds well beyond the 64-GPU regime---where
unmodeled inter-node contention would first appear---we validated two
agent-selected recipes at production scale (benchmark mode: real
kernels on 8~GPUs, restored analytically to the target).
Table~\ref{tab:scale} reports both. On DeepSeek~V3~671B (FP8,
1{,}024~GPUs / 128~nodes)---$16\times$ the largest configuration in the
sub-node sweep, and a model pairing Multi-head Latent Attention with a
3-dense$+$58-MoE layer mix seen nowhere during development---the
agent's end-to-end selection (PP=8/VPP=2/EP=8/ZBV, GBS=32{,}768)
projects 2{,}808~tok/s/GPU against 2{,}946 measured (\textbf{4.7\%}).
On Qwen3-235B-A22B (FP8, 256~GPUs / 32~nodes) the selection
(PP=4/VPP=4/EP=8/MBS=4) projects 4{,}860 vs.\ 4{,}952 measured
(\textbf{1.9\%}). Fidelity thus holds at two independent large scales,
not merely at small clean ones.

\begin{table}[h]
\centering\small\setlength{\tabcolsep}{5pt}
\begin{tabular}{llrrr}
\toprule
\textbf{Model (FP8)} & \textbf{Scale} & \textbf{Projected} & \textbf{Measured} & \textbf{Error} \\
\midrule
DeepSeek~V3 671B MoE & 1{,}024 GPU (128 node) & 2{,}808 & 2{,}946 & 4.7\% \\
Qwen3-235B-A22B MoE  & 256 GPU (32 node)      & 4{,}860 & 4{,}952 & 1.9\% \\
\bottomrule
\end{tabular}
\caption{\small Large-scale validation (tok/s/GPU). Both recipes are agent-selected end-to-end and confirmed on hardware; full search traces in Appendix~\ref{app:scale}.}
\label{tab:scale}
\end{table}

\emph{Why contention stays second-order.} In benchmark mode the
throughput-governing communication is already inside the measurement:
the heaviest collective---MoE All-to-All---is intra-node in these
recipes (EP\,$\le$\,GPUs/node) and directly executed (when EP spans
nodes it is modeled analytically instead); pipeline
P2P is a single low-volume activation per stage boundary; and DP
gradient AllReduce is issued during backward and overlapped with
compute. The residual is a small additive term, not a first-order
error---which is why the 128-node point still lands within 4.7\%. For
DeepSeek~V3 the agent also discovers the throughput-starvation insight,
scaling the global batch to the memory ceiling
(1{,}869\,$\to$\,2{,}808~tok/s/GPU); the full shape and batch sweeps
are in Appendix~\ref{app:scale}.

\subsection{\agentname Case Study: Mixtral~8$\times$22B}
\label{ssec:agent-eval}

We applied the agent (Section~\ref{sec:method-agent}) to
Mixtral~8$\times$22B (FP8, 4-node MI355X, GBS=512, seq\_len=8192),
grounded against the published AMD Mixtral~8$\times$22B 4-node BF16
reference of 3{,}475~tok/s/GPU on the same hardware.
Figure~\ref{fig:tuning-search} visualizes the search trajectory and
Table~\ref{tab:tuning-proj} reports projection-phase results;
Table~\ref{tab:tuning-real} reports real-cluster validation.

\begin{figure}[t]
  \centering
  \begin{subfigure}{0.55\linewidth}
    \centering
    \includegraphics[width=\linewidth]{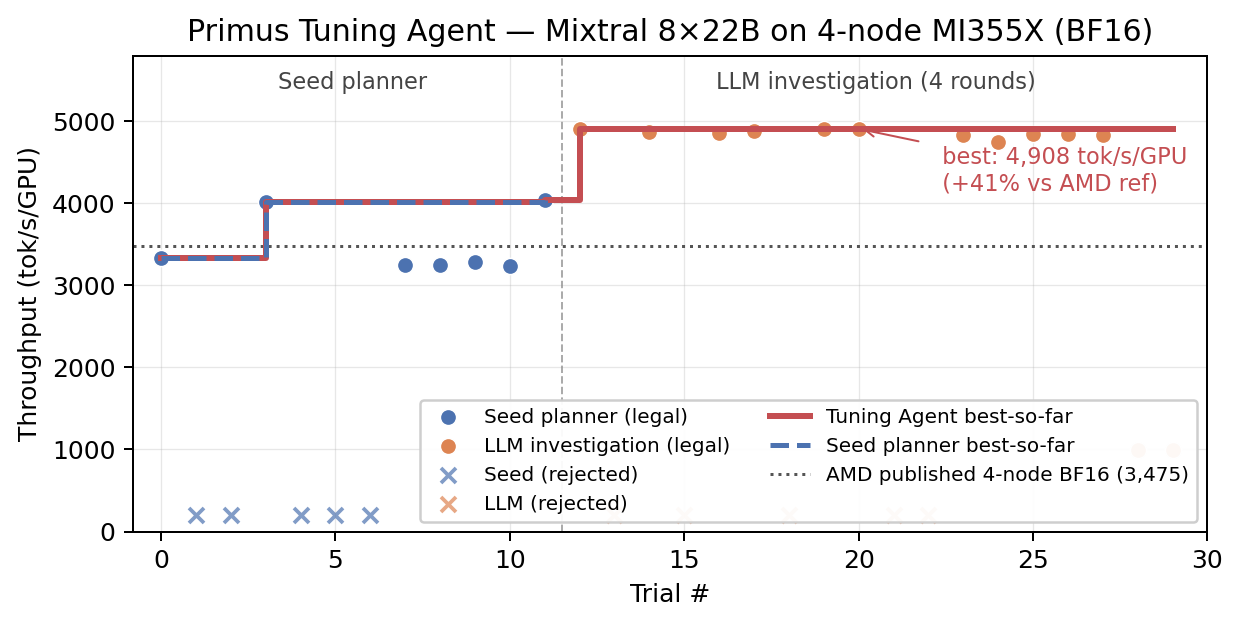}
    \caption{\small Best-so-far projected throughput per trial. The LLM jumps from the seed-planner peak (4{,}036) to 4{,}908 at its first proposal by combining MBS=1, no recompute, and DeepEP---a cross-axis combination unreachable to the seed planner alone.}
    \label{fig:tuning-convergence}
  \end{subfigure}\hfill
  \begin{subfigure}{0.42\linewidth}
    \centering
    \begin{tikzpicture}
    \begin{axis}[
      width=0.68\linewidth, height=4.3cm,
      xbar, bar width=7pt, bar shift=0pt, xmin=0, xmax=40,
      title={Memory pre-filter funnel (LLM agent run)}, title style={font=\scriptsize},
      xlabel={Trials}, label style={font=\scriptsize}, tick label style={font=\scriptsize},
      symbolic y coords={{Best configuration},{Legal trials},{Benchmarked},{Rejected by pre-filter},{Proposed configs}},
      ytick={{Best configuration},{Legal trials},{Benchmarked},{Rejected by pre-filter},{Proposed configs}},
      enlarge y limits=0.15,
      nodes near coords, every node near coord/.append style={font=\scriptsize},
    ]
    \addplot[fill=oiblue, draw=none] coordinates {(30,{Proposed configs})};
    \addplot[fill=oiblue!40, draw=none] coordinates {(7,{Rejected by pre-filter})};
    \addplot[fill=oiorange, draw=none] coordinates {(23,{Benchmarked})};
    \addplot[fill=oigreen, draw=none] coordinates {(20,{Legal trials})};
    \addplot[fill=oired, draw=none] coordinates {(1,{Best configuration})};
    \end{axis}
    \end{tikzpicture}
    \caption{\small Memory pre-filter rejects 7 out of 30 trials before any GPU benchmark, saving $\sim$175~GPU-min at zero throughput cost.}
    \label{fig:tuning-funnel}
  \end{subfigure}

  \vspace{0.5em}

  \begin{minipage}{0.46\linewidth}
    \centering\scriptsize\setlength{\tabcolsep}{2.5pt}
    \begin{tabular}{l r r r}
    \toprule
    \textbf{Mode} & \textbf{Trials} & \textbf{Best (proj.)} & \textbf{$\Delta$} \\
    \midrule
    AMD 4-node BF16 (meas.) & --  & 3{,}475 & ref. \\
    Seed planner only   & 12  & 4{,}036 & +16.1\% \\
    LLM (seed + 3)      & 30  & 4{,}908 & +41.2\% \\
    \bottomrule
    \end{tabular}
    \captionof{table}{\small Projection-phase results. LLM's +21.6\% advantage over seed planner comes from cross-axis combination (MBS=1, no recompute, DeepEP).}
    \label{tab:tuning-proj}
  \end{minipage}\hfill
  \begin{minipage}{0.50\linewidth}
    \centering\scriptsize\setlength{\tabcolsep}{2pt}
    \begin{tabular}{l l r r r}
    \toprule
    \textbf{Tag} & \textbf{Config} & \textbf{Proj.} & \textbf{Meas.} & \textbf{Err} \\
    \midrule
    llmwin12 & MBS=1 DeepEP     & 4{,}908 & \textbf{4{,}402} & $-$10.3\% \\
    llmwin19 & MBS=1 DeepEP+DO  & 4{,}903 & 4{,}377          & $-$10.7\% \\
    \bottomrule
    \end{tabular}
    \captionof{table}{\small Real-cluster validation (4-node MI355X, FP8 hybrid, 50-iter). Best measured 4{,}402~tok/s/GPU (\textbf{+27\%} over the AMD 4-node BF16 reference). DO = distributed optimizer.}
    \label{tab:tuning-real}
  \end{minipage}

  \addtocounter{figure}{-1}
  \caption{\small \agentname search on Mixtral~8$\times$22B (4-node MI355X).}
  \label{fig:tuning-search}
\end{figure}

The seed planner reaches 4{,}036~tok/s/GPU (+16.1\%) via single-axis sweeps.
The LLM jumps to a projected 4{,}908 (+41.2\% over baseline, +21.6\% over seed)
on its first proposal by combining MBS=1, no recompute, and DeepEP---a
cross-axis combination unreachable to the seed planner. The top two
non-redundant LLM proposals (\texttt{llmwin12}: DeepEP; \texttt{llmwin19}:
DeepEP + distributed optimizer) measure 4{,}402 and 4{,}377~tok/s/GPU
respectively ($-$10.3\% / $-$10.7\% vs.\ projection), delivering
\textbf{+27\%} over the AMD 4-node BF16 reference. The $\sim$10.5\%
optimistic bias is a workload-level calibration offset
(Section~\ref{ssec:rank}) that preserves
\emph{rank ordering}---the only property the agent requires. The memory
pre-filter rejected 7 of 30 trials at zero GPU cost, and the workflow
took $\sim$\textbf{30~minutes} of single-node exploration.

\subsection{Rank Preservation and the Calibration Offset}
\label{ssec:rank}

Search requires only that the projected \emph{ordering} of candidates
match the measured ordering---not absolute throughput. We test this
directly with paired projected-vs-measured sweeps on three production
MoE models, each spanning the axes the agent moves (recompute, VPP, PP,
MBS, GBS); Table~\ref{tab:rank} summarizes and
Appendix~\ref{app:rank} gives every per-config pair for all three models.

\begin{table}[h]
\centering\small\setlength{\tabcolsep}{5pt}
\begin{tabular}{lccccc}
\toprule
\textbf{Workload} & \textbf{$n$} & \textbf{Spearman $\rho$} & \textbf{Kendall $\tau$} & \textbf{Offset $k$ (cv)} & \textbf{Regret} \\
\midrule
Mixtral-8$\times$22B FP8 (32 GPU) & 7 & 1.000 & 1.000 & 1.128 (2.4\%)  & 0\% \\
Qwen3-235B-A22B FP8 (32 GPU)      & 7 & 1.000 & 1.000 & 0.901 (11.3\%) & 0\% \\
GPT-OSS-120B BF16 (64 GPU)        & 9 & 0.933 & 0.833 & 1.121 (5.4\%)  & 0\% \\
\bottomrule
\end{tabular}
\caption{\small Rank preservation across three MoE models spanning FP8/BF16 and 32--64 GPUs. Selection regret (agent pick vs.\ measured-optimal) is 0\% throughout.}
\label{tab:rank}
\end{table}

Selection regret is 0\% on all three---the agent always picks the
measured-optimal configuration. The offset is not an exactly constant
term but a workload-level \emph{multiplicative} factor (isolated-kernel
benchmark vs.\ sustained run) that drifts mildly with recompute depth;
crucially, Mixtral ($k$=1.13, optimistic) and Qwen3 ($k$=0.90,
pessimistic) carry \emph{opposite-sign} offsets, yet both reproduce the
ordering exactly, so rank preservation is not offset-dependent. Order
survives because the drift is small relative to the throughput gaps
between configurations; individual off-optimal configurations in these
sweeps can deviate by more than 10\% without affecting selection.
On GPT-OSS-120B ($\rho$=0.933) the three
discordant pairs (of 36) are confined to a near-tied mid-pack band
within $\sim$9\% of one another; every pair separated by
$>$\,$\sim$9\% measured throughput is ordered correctly and the top-4
and bottom-2 are exact (Appendix~\ref{app:rank}). Note that
the benchmark oracle must run on an \emph{idle} node---co-located load
inflates kernel times and depresses projections; a clean run restores
$\rho$=1.000.

\subsection{Comparison with Analytical and Auto-Parallelism Baselines}
\label{ssec:baselines}

\textbf{vs.\ Calculon.} On Llama-3.1-70B measured on AMD MI325X
(8~nodes, 64~GPUs) Primus averages
\textbf{6.6\%} absolute error against Calculon's~\cite{calculon}
\textbf{28.9\%}; Calculon's opposite-direction errors ($+$16.2\% FP8,
$-$41.6\% BF16) inflate the measured 1.5$\times$ FP8--BF16 gap to
2.9$\times$, distorting any precision decision that relies on it. By
construction, it has no experts, All-to-All, or ROCm software path, so Mixtral
and DeepSeek~V3 are inexpressible (Appendix~\ref{app:baselines}).

\textbf{vs.\ Galvatron-style search.} We implement a DP-style
enumeration in the spirit of Galvatron~\cite{galvatron} over 530
configurations and hand it our own (more expressive) simulate
oracle (an advantage to the baseline). On Mixtral (4-node
MI355X, FP8) an exhaustive sweep of all 530 configs reaches only
3{,}656~tok/s/GPU vs.\ \textbf{4{,}698} for \agentname{}$+$Benchmark
(Table~\ref{tab:ablation}). The gap is the \emph{oracle}, not the
search: the benchmark directly measures superposed FP8 tensor-core,
ROCm Communication Collectives Library (RCCL) A2A, and grouped-GEMM effects that any analytical model
approximates, and correctly ranks the winning intra-node
EP=8/DeepEP topology. An ablation isolates the two contributions: the
benchmark oracle adds \textbf{+36\%} over the simulate oracle at equal
agent, and the agent framework adds \textbf{+10\%} over a Claude Code
agent (Claude Opus~4.6, no agent framework) with the same oracle.
The Galvatron-style search has no precision knob, so unlike our agent
it cannot switch from BF16 to FP8 mid-search (at a 30-trial budget it
stays BF16, 2{,}155); with FP8 fixed up front it reaches the 3{,}656
above, so precision does not explain the gap.

\begin{table}[h]
\centering\small\setlength{\tabcolsep}{5pt}
\begin{tabular}{llrr}
\toprule
\textbf{Method} & \textbf{Oracle} & \textbf{Proj.\ tok/s/GPU} & \textbf{vs.\ baseline} \\
\midrule
\agentname + Benchmark         & Benchmark & \textbf{4{,}698} & +39.5\% \\
Claude Code + Benchmark        & Benchmark & 4{,}275 & +26.9\% \\
Galvatron-style FP8 exhaustive & Simulate  & 3{,}656 & +8.6\% \\
\agentname + Simulate          & Simulate  & 3{,}445 & +2.3\% \\
Galvatron-style budget=30      & Simulate  & 2{,}155 & $-$36.0\% \\
\bottomrule
\end{tabular}
\caption{\small Ablation and baseline comparison on Mixtral-8$\times$22B (4-node MI355X, FP8; baseline 3{,}368 projected tok/s/GPU). Oracle contribution $+$36\%, agent-framework contribution $+$10\%; winning configs in Table~\ref{tab:ablation-configs}. Run on a different 4-node cluster from the case study in Section~\ref{ssec:agent-eval} (projected 4{,}908); cluster-to-cluster variation is $<$5\%. Details in Appendix~\ref{app:baselines}.}
\label{tab:ablation}
\end{table}

\subsection{MoE Load-Balancing Robustness}
\label{ssec:moe-balance}

\emph{Benchmark runs are balanced by construction.} We force load
balancing for benchmarking only\footnote{Megatron flag
\texttt{moe\_router\_force\_load\_balancing}; never used in real training.}:
a random straight-through router plus an exactly even dispatcher make
executed load uniform, so the imbalance
$I=$~max/mean is 1 and the uniform-load assumption holds exactly.
Real training sits close to this: aux-loss (or aux-loss-free)
balancing plus large-batch averaging, under which per-expert deviation
shrinks as $1/\sqrt{\text{tokens/expert}}$, gives a steady-state
$I\approx1.1$--1.3.
\emph{When balancing is off}, grouped-GEMM time is gated by the busiest
EP-rank rather than the mean, and the A2A jumps regime. Sweeping $I$
from 1$\times$ to 4$\times$ on real MI355X ($I\ge2$ is a deliberate
stress test beyond the training regime), a busiest-rank correction
(compute gated by $I$ at experts-per-rank $E/\text{EP}$, plus an A2A
regime term) cuts worst-case error from 53--71\% to under 10\%.
\emph{On a real trained model} (DeepSeek~V3 671B, native learned
balancing) the per-layer imbalance has median 1.38 at EP=8 (2.13 at
EP=32), but per-layer spikes fall on different devices, so the
step-level \emph{pooled} imbalance that gates throughput is only
1.05--1.12; at EP=8 the corrected error is $\approx$2\%.
\emph{Across EP}, the penalty depends only on $E/\text{EP}$, so a fit at
one EP predicts the others (Mixtral, EP$\in\{2,4,8\}$, $\pm$6\%). This
is what lets us calibrate on the 8-GPU benchmark and project to the
128-node target (Appendix~\ref{app:moe}).

\subsection{Throughput Improvement}
\label{ssec:throughput-improvement}

On Mixtral~8$\times$22B (4-node MI355X), the agent delivers \textbf{+27\%}
measured throughput over the AMD BF16 reference, with $\sim$25~GPU-min per
accepted trial and 7 of 30 trials rejected at zero cost by the memory
pre-filter (Section~\ref{ssec:agent-eval}). At production scale the same
workflow selects the measured-optimal recipe end-to-end on
DeepSeek~V3~671B (1{,}024~GPU, 4.7\% error) and Qwen3-235B-A22B
(256~GPU, 1.9\% error), confirming the gains hold well beyond the
sub-node regime (Section~\ref{ssec:scale}).

\subsection{Scope and Limitations}
\label{ssec:limits}

\sysname is validated on AMD Instinct GPUs and the ROCm software stack, but
nothing in the modeling is vendor-specific: hardware profiles are YAML
(peak FLOPS, HBM, collective topology) and the microbenchmark runs on
any supported GPU ($\sim$8~min on 8 GPUs). We have \emph{not} yet
empirically validated on NVIDIA GPUs or tensor processing units (TPUs);
porting to NVIDIA GPUs requires only a hardware YAML and the 8-GPU
benchmark (TPUs also need an Accelerated Linear Algebra (XLA)
collective model). Network contention is not modeled; it stayed
second-order in our evaluations (Section~\ref{ssec:scale}) but could
add error in some inter-node EP or sub-node benchmark settings. ZeRO-2 is excluded, and FP8 recipes must be
convergence-validated before production. Appendix~\ref{app:limitations}
covers these, and extensions to new layer types.

\section{Related Work}
\label{sec:related}

\textbf{Analytical performance models.}
Roofline analysis~\cite{roofline} provides an upper bound on kernel
throughput from compute and memory-bandwidth peaks, but ignores fused
kernels, pipeline schedules, and multi-dimensional parallelism
interactions. Calculon~\cite{calculon} and Galvatron~\cite{galvatron}
extend this with closed-form analytical models that jointly search
parallelism configurations (TP, PP, DP) in milliseconds, enabling
co-design sweeps (Galvatron-BMW~\cite{galvatronbmw} adds memory
balancing). Calculon uses only hardware specifications; Galvatron
profiles single-device layer compute but models communication from
nominal bandwidth. Neither captures clock-throttling under sustained
load, kernel-library drift, or grouped-GEMM wave-quantization in
MoE---effects benchmark anchoring resolves automatically.
FlexFlow~\cite{flexflow} and Alpa~\cite{alpa} require operator- or
stage-level profiling on the target cluster.

\textbf{Trace-driven and simulation-based approaches.}
Trace-driven simulators such as ASTRA-sim~\cite{astrasim} and
Echo/Moye~\cite{echo} achieve higher fidelity by replaying workload traces
but require target-hardware execution traces or extensive calibration;
Lumos~\cite{lumos} proposes a toolkit that reaches 3.3\% average error
on GPT-3 variants on clusters of up to 512 H100 GPUs using trace
replay, while ProTrain~\cite{protrain} builds high-fidelity cost models
from a runtime profiler to automate memory-management policies,
achieving 1.43--2.71$\times$ throughput improvements.
Proteus~\cite{proteus} simulates hybrid-parallel strategies with
compute--communication overlap but targets CUDA and does not model
expert-parallel All-to-All; Vidur~\cite{vidur} simulates LLM
inference only. Optimus~\cite{optimus}
fits online resource-performance models to estimate training speed as a
function of allocated resources, enabling dynamic scheduling in deep
learning clusters. These systems are oriented toward profiling and
scheduling existing target-cluster runs; in contrast, \sysname's
pure-simulation mode runs on CPU and supports pre-silicon design-space
exploration without any execution trace from the target, and its hybrid
mode requires only a single-node benchmark to project multi-node
throughput.

\textbf{Hyperparameter optimization: from classical search to LLM agents.}
The Tree-structured Parzen Estimator (TPE)~\cite{bergstra2011} and
Gaussian-process Bayesian optimization (BO)~\cite{snoek2012} showed that surrogate-guided search beats grid
search. BO fits LLM-training configuration poorly: it assumes a
stationary objective, whereas a 4-byte activation miscount can toggle
legality; it treats every trial as an equal-cost black box, whereas many
candidates can be rejected analytically before any GPU time; and it
ignores cross-axis structure (e.g., the MoE backend interacts only with
EP knobs). LLM agents now replace or augment BO: AgentHPO~\cite{agenthpo}
pairs a Creator LLM with an Executor that trains and reports feedback,
and LLAMBO~\cite{llambo} frames BO candidate proposal in natural
language. Both target model-accuracy objectives; our problem adds a
combinatorial feasibility filter and a multi-level surrogate
(analytical memory check, then benchmark projection) that neither
exploits.

\textbf{Configuration tuning for distributed training.}
DeepSpeed Autotuning~\cite{deepspeed-autotune} grid-searches ZeRO
stage and micro-batch on the target cluster but does not consider the
joint parallelism shape or the pipeline schedule; FlexFlow~\cite{flexflow},
Alpa~\cite{alpa}, and Galvatron~\cite{galvatron} explore parallelism
plans with their own cost models but, like the analytical baselines
above, lack a benchmark-anchored projection step and do not use an
LLM reasoning loop to propose cross-axis combinations. The \agentname is the first, to our knowledge, to couple a
benchmark-anchored projection oracle (seconds per candidate, no
multi-node cluster required) with an LLM-driven cross-axis
investigation loop implemented in DSPy~\cite{dspy} in the spirit of
ReAct~\cite{react}-style reasoning and recursive tool
use~\cite{rlm}.

\section{Conclusion}
\label{sec:conclusion}

We presented \sysname, a hybrid framework for memory and performance
projection of distributed training that benchmarks compute and in-scope
communication on as few as one device and analytically reconstructs
out-of-scope communication and the full pipeline schedule at the target
scale. The \agentname uses it as a fast surrogate to search the joint
configuration space end-to-end. Across dense and MoE models on MI325X
and MI355X, projections of the validated recipes stay within $\sim$10\%
of measured throughput (4.7\% at 1{,}024~GPUs), and the agent delivers
+27\% over the AMD Mixtral 4-node BF16 reference in $\sim$30~minutes of
single-node exploration, finding cross-axis combinations no
deterministic planner reaches, without multi-node hardware.

\begin{ack}
This work was funded by Advanced Micro Devices, Inc. All authors are
employees of AMD. The authors declare no other competing interests.
\copyright{} 2026 Advanced Micro Devices, Inc. All rights reserved. AMD,
the AMD Arrow logo, AMD Instinct, ROCm, and combinations thereof are
trademarks of Advanced Micro Devices, Inc. Other product names used in
this publication are for identification purposes only and may be
trademarks of their respective owners.
\end{ack}

\clearpage
\appendix

\titleformat{\section}[block]{\Large\bfseries}{Appendix \thesection:}{0.5em}{}
\titlespacing*{\section}{0pt}{1.5ex plus 0.5ex}{1.0ex plus 0.3ex}

\begin{center}
{\LARGE\bfseries Appendices}
\end{center}
\vspace{1em}

\section{Hardware and Software Environment}
\label{app:env}

All measurements reported in this paper were obtained on AMD Instinct
MI300X, MI325X, and MI355X GPUs running the ROCm software stack with the
hipBLASLt, Composable Kernel, and Triton kernel libraries. Distributed
training uses the Megatron-LM~\cite{megatron} backend through the Primus
framework. Hardware profiles for the GPUs used in this paper are summarized
in Table~\ref{tab:gpus}; custom profiles for arbitrary cluster topologies
and hypothetical pre-silicon architectures can be added via YAML.
\sysname and the \agentname are open source in the AMD Primus framework
at \url{https://github.com/AMD-AGI/Primus} (MIT license).

\begin{table}[h]
\centering
\footnotesize
\setlength{\tabcolsep}{4pt}
\begin{tabular}{llrrr}
\toprule
\textbf{GPU} & \textbf{Arch} & \textbf{HBM (GB/s)} & \textbf{CUs} & \textbf{XCDs} \\
\midrule
MI300X & gfx942 & 5,300 & 304 & 8 \\
MI325X & gfx942 & 6,000 & 304 & 8 \\
MI355X & gfx950 & 8,000 & 256 & 8 \\
\bottomrule
\end{tabular}
\caption{\small AMD Instinct GPU architectures used in this paper. CUs = compute units; XCDs = accelerator complex dies.}
\label{tab:gpus}
\end{table}

\paragraph{Two practical consequences of this stack.}
First, every ROCm software release shifts achievable kernel performance, so a pure
analytical model would need continual recorrelation; a benchmark-anchored
approach reflects the kernels installed on the system at the time the
projection is run. Second, GPU clock frequency on these parts varies
between attention, dense MLP, and MoE expert GEMMs because of power-cap
throttling under sustained dense matrix workloads, with a 44--70\% gap
between peak and maximum achievable FLOPS on previous-generation AMD hardware~\cite{amd-maf}.
Benchmarking captures both effects without requiring per-kernel
recorrelation.

\section{Analytical Models}
\label{app:models}

This appendix details the analytical models used by \sysname's projection
oracle, organized by component.

\subsection{Memory Projection}
\label{app:models-mem}

Memory is estimated using a hierarchical profiler tree that mirrors the
model's module structure. Each profiler implements two methods: one for
parameter counts (total or per-GPU with sharding) and one for activation
memory per microbatch.

\paragraph{Parameter and optimizer memory:}
Per-GPU parameter count accounts for parallelism-induced sharding: TP
splits weight matrices; PP assigns layers to stages; EP shards expert
parameters while replicating non-expert parameters; CP replicates all
parameters; and FSDP/ZeRO shards across DP. With per-parameter byte
costs $B_w$ (weight), $B_g$ (gradient), and $B_o$ (optimizer state):
\begin{equation}
\text{bytes\_per\_param} = B_w + B_g + B_o/\text{DP}
\end{equation}
For BF16 mixed-precision with Adam, $(B_w, B_g, B_o) = (2, 2, 10)$
(2~B BF16 weight, 2~B BF16/FP32 gradient, 10~B FP32 optimizer state =
4~B master weight + 4~B first moment + 2~B second moment shared) so
$\text{bytes\_per\_param} = 4 + 10/\text{DP}$. Under FP8 mixed precision
the weight term drops to 1~B; the distributed-optimizer flag further
shards $B_o$ across DP ranks. The projector reads these parameters
from the framework configuration, so a single equation covers all
combinations.

\paragraph{Activation memory:}
Activation memory is parameterized by a saved-activation byte cost
$B_{\text{act}}$ that is read from the framework's recompute/save policy
($B_{\text{act}}=2$ for BF16, $B_{\text{act}}=1$ for FP8 saved
activations). The MoE MLP activation, the dominant contributor, is:
\begin{equation}
\begin{split}
\text{act\_moe} = {} & \text{tokens} \times (\text{topk} + N_{\text{shared}}) \\
                     & \times (H + 3 \times \text{FFN}_e) \times B_{\text{act}}
\end{split}
\end{equation}
For a large MoE model (MBS=4, S=16384, CP=4, H=8192, FFN$_e$=2048,
topk=36) at BF16 saved activations, a single MoE layer reaches
\textbf{16.19~GB}---over $30\times$ the attention activation
($\sim$0.51~GB)---and roughly halves to $\sim$8.1~GB if the framework
saves activations in FP8. Attention, MLP, layer-norm, residual,
embedding, output, and (non-fused) loss profilers each apply
$B_{\text{act}}$ to their tensor shapes; the loss profiler additionally
allocates an FP32 (4~B) softmax scratch when CE loss fusion is off.

\paragraph{Pipeline schedule scaling:}
$\text{total\_act} = \text{base}\times\text{PP}\times F_{\text{intlv}}
\times F_{\text{ga}}$ where
$F_{\text{intlv}} = 1 + (\text{PP}-1)/(\text{PP}\cdot\text{VPP})$ is the
interleaved penalty and $F_{\text{ga}} = \min(1, \text{GA}/\text{PP})$
handles cases where gradient accumulation steps are fewer than pipeline
stages.

\paragraph{Recomputation:}
With full recomputation, only the layer input tensor is checkpointed
($\text{MBS}\times S/(\text{TP}\cdot\text{CP})\times H\times B_{\text{act}}$),
yielding the same precision-aware reduction as the activation
formula---e.g., a $\sim$72$\times$ shrink from 18.20~GB per MoE layer
to $\sim$0.25~GB at BF16, or to $\sim$0.13~GB at FP8. Partial (selective)
recomputation is also supported and inherits $B_{\text{act}}$ on the
checkpointed subset.

\subsection{Performance Projection (Sub-Node Methodology)}
\label{app:models-perf}

The body (Section~\ref{sec:subnode}) describes the three phases of
the sub-node workflow---downscale, measure, upscale---and the
priority order $\text{PP}\to\text{EP}\to\text{TP}$. This appendix
records two operational details. \emph{Profiling modes:}
\texttt{benchmark} (default, requires GPU) measures real kernels;
\texttt{simulate} (no GPU) uses Origami~\cite{origami} GEMM and a
Flash Attention~v3~\cite{flashattn3} tile-level simulator on CPU;
\texttt{both} runs side-by-side for cross-validation
(Table~\ref{tab:bench-vs-sim}). \emph{Restoration formulas:} EP is
restored per MoE layer by adding the A2A delta $\Delta =
\text{A2A}(C^{*}) - \text{A2A}(C^{\text{bench}})$, where both terms
are evaluated by the collective model below for the target and the
benchmarked EP group, respectively, with the measured intra-node A2A
used as an anchor when decomposed timing is available. TP applies two adjustments: compute is rescaled by
$\text{bench\_tp}/\text{target\_tp}$, and the AllReduce contribution
is replaced via the ratio form $\text{target}_{\text{AR}} =
\text{measured}_{\text{AR}}\times
[\text{analytical}(\text{target})/\text{analytical}(\text{bench})]$
  when measured AllReduce data is available, falling back to the pure
analytical model otherwise. PP is restored by re-simulating the
target schedule from the benchmarked per-layer times.

\begin{table}[h]
\centering
\scriptsize
\setlength{\tabcolsep}{3pt}
\begin{tabular}{p{1.5cm} p{2.5cm} p{2.5cm}}
\toprule
\textbf{Factor} & \textbf{Analytical Sim.} & \textbf{GPU Benchmark} \\
\midrule
Hardware       & No GPU required          & Requires target GPUs \\
Speed          & Fast, deterministic      & Slower, with variance \\
Pre-silicon    & Primary use case         & Not possible \\
GPU freq.      & Estimated from model     & Captures actual throttling \\
SW stack       & Needs recorrelation      & Reflects current kernels \\
Grouped GEMM   & Depends on model fidelity & Measured directly \\
Framework      & Not captured             & Included \\
Comm.          & Analytical for all       & Measured when in scope \\
Pipeline       & Simulated exactly        & Simulated exactly \\
\bottomrule
\end{tabular}
\caption{\small Benchmark vs.~simulation trade-offs.}
\label{tab:bench-vs-sim}
\end{table}

\subsection{Communication Modeling}
\label{app:models-comm}

\paragraph{Topology domains:} The communication model distinguishes three
topology domains: \emph{intra-node} ($\sim$1{,}024~GB/s via xGMI,
0.45~$\mu$s latency), \emph{inter-node} ($\sim$50~GB/s per network interface card (NIC) via InfiniBand (IB) or
remote direct memory access over Converged Ethernet (RoCE), 2.0~$\mu$s), and \emph{inter-pod} ($\sim$25~GB/s,
10.0~$\mu$s).

\paragraph{Algorithm selection:} For each collective, the model evaluates
all applicable algorithms and selects the minimum-time configuration.
\textbf{AllReduce}: Ring, Hypercube, Bruck, and Single-shot algorithms
$\times$ four protocols---up to 16 candidates. \textbf{All-to-All}:
Direct, Single-shot, Hierarchical, and PXN/DeepEP. \textbf{P2P
Send/Recv}: latency~+~bandwidth model.

\paragraph{MoE-aware gradient AllReduce:} For MoE models with EP$>$1,
gradient AllReduce is decomposed into expert gradient AR (across DP/EP
replicas, inter-node) and non-expert gradient AR (across full DP group).
When EP$>$1, A2A synchronization barriers prevent gradient AR from
overlapping with backward compute; the full AR time is placed on the
critical path.

\subsection{FSDP Overlap Model}
\label{app:models-fsdp}

FSDP shards parameters with three collectives: a forward AllGather that
prefetches sharded parameters, a recompute AllGather that re-materializes
parameters during backward when activation recomputation is on, and a
ReduceScatter that aggregates gradients at the end of backward. A
per-layer overlap model determines what fraction of each collective hides
behind adjacent-layer compute, scaled by the per-layer compute/comm ratio
and clamped by phase-specific ceilings:
\begin{equation}
\text{overlap}(\text{comm}, \text{compute}, \text{ceiling}) = \min\Bigl(1, \frac{\text{compute}}{\text{comm}}\Bigr) \cdot \text{ceiling}
\end{equation}
with ceilings 0.95 for forward AllGather prefetch, 0.93 for recompute
AllGather, and 0.92 for ReduceScatter. Only the exposed portion
$\text{comm}\cdot(1-\text{overlap})$ enters the critical path.

\subsection{Pipeline Schedule Algorithms}
\label{app:models-pipe}

The pipeline simulator supports six scheduling algorithms
(Table~\ref{tab:pipeline}). Zero-bubble scheduling~\cite{zerobubble}
exploits the observation that the weight gradient (W) has no data
dependency on the next pipeline stage---only the input gradient (B) does.
Since W can be scheduled flexibly, it fills pipeline bubbles. For VPP=1,
the basic Zero-Bubble scheduler uses a fixed F-B-W steady-state pattern,
while Sea AI Lab's ILP-based scheduler~\cite{zerobubble} optimally places
W operations using integer linear programming. For VPP=2, the ZBV
(Zero-Bubble V-shape) family extends this across two virtual pipeline
chunks: \textbf{ZBV Formatted} uses deterministic phase transitions,
while \textbf{ZBV Greedy} uses a greedy placement algorithm with
configurable memory modes.

\begin{table}[h]
\centering
\scriptsize
\setlength{\tabcolsep}{2pt}
\begin{tabular}{lcp{3.8cm}}
\toprule
\textbf{Algorithm} & \textbf{VPP} & \textbf{Description} \\
\midrule
1F1B              & 1   & Standard one-forward-one-backward \\
Interleaved 1F1B~\cite{megatron-v2}  & $>$1 & Multiple virtual chunks per rank; reduces bubble by VPP factor \\
Zero-Bubble       & 1   & Splits backward into B+W; F-B-W steady-state fills bubbles \\
ZBV Formatted     & 2   & V-shape schedule with structured warm-up / stable / cool-down phases \\
ZBV Greedy        & 2   & V-shape with greedy placement and memory-aware modes \\
Sea AI Lab ILP    & 1   & ILP-optimal W placement~\cite{zerobubble} \\
\bottomrule
\end{tabular}
\caption{\small Supported pipeline scheduling algorithms.}
\label{tab:pipeline}
\end{table}

\section{\agentname Implementation}
\label{app:agent}

\subsection{Seed Planner Axis Schedule}
\label{app:agent-seed}

The deterministic seed planner walks a fixed schedule of single-axis
sweeps anchored on the user-supplied baseline. Axes are visited in the
following order, each producing a small bounded number of candidate
configurations:
\begin{enumerate}
\item \textbf{Memory frontier}: full $\to$ selective $\to$ no
  recomputation, holding all other axes fixed.
\item \textbf{MoE communication backend}: standard A2A vs.\ DeepEP
  (only emitted when the model is MoE and EP$>$1).
\item \textbf{Pipeline bubble reduction}: 1F1B baseline,
  Interleaved~1F1B with VPP$\in\{2,3,4\}$, Zero-Bubble, and Sea AI Lab
  ILP, restricted to schedules that fit the per-layer memory budget.
\item \textbf{Second-order axes}: micro-batch size
  $\in\{\text{baseline}, 1, 2\}$ and the distributed-optimizer flag.
\end{enumerate}
For each candidate, the analytic memory pre-filter rejects configurations
projected above the cap before any GPU time is consumed. Surviving
candidates are scored by the projection oracle and recorded in a shared
trial ledger that the LLM phase reads on entry.

\subsection{LLM Investigation Loop}
\label{app:agent-llm}

The LLM phase uses a two-LLM design built on DSPy~\cite{dspy}, served
through an OpenAI-compatible LiteLLM gateway. The \emph{Planner} is
Claude~Sonnet~4.6 (\texttt{claude-sonnet-4-6}, ChainOfThought) and the
\emph{Recursive Language Model} (RLM)~\cite{rlm} is Claude~Opus~4.6 (\texttt{claude-opus-4-6}, ReAct
tool loop); we set \texttt{max\_tokens}=8000, \texttt{timeout}=300\,s,
$\le$25 candidates/round and $\le$4 rounds, with temperature/top-p at
DSPy defaults. Verbatim prompts are reproduced in
Appendix~\ref{app:prompts}.
\begin{itemize}
\item \textbf{Planner LLM} reads the seed-phase trial ledger
  (configuration, projected throughput, projected memory, legality, and
  failure mode where applicable) and emits a hypothesis paragraph
  together with a JSON list of candidate configurations to evaluate next.
\item \textbf{RLM} executes those candidates one at a time,
  calling the projection oracle as a tool, writing notes to a per-round
  scratchpad, and optionally querying a sub-LLM for deeper analysis of
  intermediate results.
\end{itemize}
Each round runs up to 25 candidate evaluations; we cap the agent at four
rounds, after which the planner is re-prompted with the cumulative ledger.
The LLM is permitted to cross-product axes the seed planner treats
independently (e.g., recomputation $\times$ MoE backend $\times$ pipeline
schedule), which is the source of the +21.6\% advantage over the seed
plan reported in Section~\ref{ssec:agent-eval}.

\subsection{Memory Pre-Filter and Trial Selection}
\label{app:agent-filter}

A trial is \textbf{legal} iff (i) axis validation passes
(PP$\times$DP$\times$CP$\times$TP tiles the target cluster), (ii) the
projected per-GPU memory satisfies
$\text{mem} \le \text{cap} = (\text{HBM} - \text{system overhead})\times 0.9$,
and (iii) the benchmark returns a parseable throughput. The overall
winner is $\arg\max(\text{legal},\text{key}=\text{tok/s/GPU})$. The cap
calibration is the basis for the memory pre-filter of
Section~\ref{ssec:agent-eval}, and its predictions agreed perfectly with
hardware: every configuration projecting
$\le$~cap ran successfully on real hardware, every configuration projecting
$>$~cap OOMed on the real cluster.

\section{Large-Scale Search Traces}
\label{app:scale}

This appendix expands the large-scale validation of
Section~\ref{ssec:scale}. Both runs use benchmark mode (real kernels on
8~GPUs) with analytical restoration to the multi-node target.

\paragraph{DeepSeek~V3 671B, 1{,}024 GPUs (128 nodes), FP8.}
The agent first sweeps the pipeline shape at fixed GBS=4{,}096, then
scales the global batch on the winning shape. Tables~\ref{tab:ds-shape}
and~\ref{tab:ds-batch} show the trajectory; the final selection is
validated against a measured run at equal GBS=32{,}768
(Table~\ref{tab:ds-val}).

\begin{table}[h]
\centering
\begin{minipage}[t]{0.31\linewidth}
\centering\scriptsize\setlength{\tabcolsep}{3pt}
\captionof{table}{\small Shape sweep (GBS=4{,}096).}
\label{tab:ds-shape}
\begin{tabular}{lr}
\toprule
\textbf{Shape} & \textbf{tok/s} \\
\midrule
PP2/VPP2/1F1B          & 1{,}017 \\
PP4/VPP2/1F1B          & 1{,}408 \\
PP8/VPP2/1F1B          & 1{,}752 \\
PP8/VPP1/zero-bubble   & 1{,}852 \\
\textbf{PP8/VPP2/ZBV}  & \textbf{1{,}869} \\
\bottomrule
\end{tabular}
\end{minipage}\hfill
\begin{minipage}[t]{0.33\linewidth}
\centering\scriptsize\setlength{\tabcolsep}{3pt}
\captionof{table}{\small Batch scaling (PP8/ZBV).}
\label{tab:ds-batch}
\begin{tabular}{llr}
\toprule
\textbf{MBS} & \textbf{GBS} & \textbf{tok/s} \\
\midrule
2 & 4{,}096  & 1{,}869 \\
2 & 16{,}384 & 2{,}424 \\
8 & 16{,}384 & 2{,}724 \\
\textbf{8} & \textbf{32{,}768} & \textbf{2{,}808} \\
\bottomrule
\end{tabular}
\end{minipage}\hfill
\begin{minipage}[t]{0.33\linewidth}
\centering\scriptsize\setlength{\tabcolsep}{3pt}
\captionof{table}{\small Validation (GBS=32{,}768).}
\label{tab:ds-val}
\begin{tabular}{lr}
\toprule
\textbf{Quantity} & \textbf{tok/s} \\
\midrule
Measured (sel.\ recompute) & 2{,}946 \\
Agent projection           & 2{,}808 \\
Error                      & 4.7\% \\
\bottomrule
\end{tabular}
\end{minipage}
\end{table}

The agent identifies the throughput-starvation insight---batch to the
memory ceiling---and lifts throughput from 1{,}869 to
2{,}808~tok/s/GPU, then the projection lands within 4.7\% of the
measured 2{,}946 at 128 nodes.

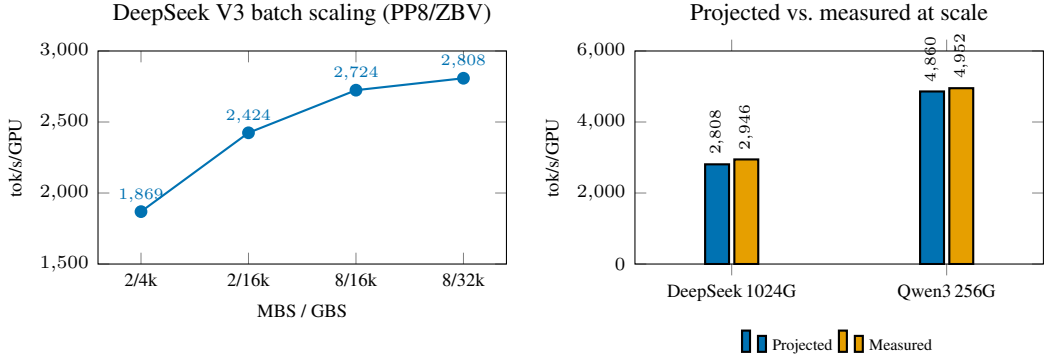
\begin{figure}[h]
\centering
\begin{tikzpicture}
\begin{groupplot}[
  group style={group size=2 by 1, horizontal sep=1.7cm},
  width=0.5\linewidth, height=4.4cm,
  ymin=0, tick label style={font=\scriptsize},
  label style={font=\scriptsize}, title style={font=\small},
  every axis plot/.append style={thick},
]
\nextgroupplot[
  title={DeepSeek~V3 batch scaling (PP8/ZBV)},
  ylabel={tok/s/GPU}, ymin=1500, ymax=3000,
  xtick={1,2,3,4},
  xticklabels={{2/4k},{2/16k},{8/16k},{8/32k}},
  xlabel={MBS / GBS}, xmin=0.6, xmax=4.4,
  nodes near coords, every node near coord/.append style={font=\scriptsize},
]
\addplot[oiblue, mark=*, mark size=2pt] coordinates {(1,1869) (2,2424) (3,2724) (4,2808)};
\nextgroupplot[
  title={Projected vs.\ measured at scale},
  ylabel={tok/s/GPU}, ymin=0, ymax=6000,
  ybar, bar width=9pt,
  symbolic x coords={DeepSeek\,1024G, Qwen3\,256G},
  xtick=data, x tick label style={font=\scriptsize},
  enlarge x limits=0.45,
  legend style={font=\scriptsize, at={(0.5,-0.28)}, anchor=north, legend columns=2, draw=none},
  nodes near coords, every node near coord/.append style={font=\scriptsize, rotate=90, anchor=west},
]
\addplot[fill=oiblue] coordinates {(DeepSeek\,1024G,2808) (Qwen3\,256G,4860)};
\addplot[fill=oiorange] coordinates {(DeepSeek\,1024G,2946) (Qwen3\,256G,4952)};
\legend{Projected, Measured}
\end{groupplot}
\end{tikzpicture}
\caption{\small \textbf{Left:} the agent's batch-scaling trajectory on
DeepSeek~V3 671B at 1{,}024~GPUs---batching to the memory ceiling lifts
throughput 50\% (1{,}869$\to$2{,}808~tok/s/GPU). \textbf{Right:}
projected vs.\ measured throughput at production scale; error is 4.7\%
(DeepSeek~V3, 1{,}024~GPU) and 1.9\% (Qwen3-235B, 256~GPU).}
\label{fig:scale}
\end{figure}

\paragraph{Qwen3-235B-A22B, 256 GPUs (32 nodes), FP8.}
Independently, the agent selected TP=1/PP=4/EP=8/CP=1/VPP=4/DP=64,
MBS=4, GBS=8{,}192, seq=4{,}096, projecting 4{,}860 vs.\ 4{,}952
measured (1.9\%), ranked first over a deeper PP=16 alternative. (The
32-GPU rank-sweep of Appendix~\ref{app:rank} uses the related VPP=3
shape; the 256-GPU run uses VPP=4.)

\section{Rank Preservation: Full Per-Config Tables}
\label{app:rank}

This appendix gives every projected-vs-measured pair behind
Table~\ref{tab:rank}. All three sweeps are \emph{fully} measured: every
listed configuration was run on hardware, so the ordering comparison is
exact.

\begin{table}[h]
\centering\small\setlength{\tabcolsep}{5pt}
\begin{tabular}{lrcrrr}
\toprule
\textbf{Tag} & \textbf{Recomp.\ layers} & \textbf{MBS} & \textbf{Projected} & \textbf{Measured} & \textbf{$k$} \\
\midrule
r00\_b1          & 0  & 1 & 4{,}957 & 4{,}213 & 1.177 \\
r03\_b1          & 3  & 1 & 4{,}620 & 4{,}059 & 1.138 \\
r09\_b2          & 9  & 2 & 4{,}397 & 3{,}884 & 1.132 \\
r06\_b1          & 6  & 1 & 4{,}353 & 3{,}860 & 1.128 \\
r09\_b1          & 9  & 1 & 4{,}095 & 3{,}709 & 1.104 \\
r14\_b2          & 14 & 2 & 4{,}029 & 3{,}558 & 1.132 \\
r14\_b1          & 14 & 1 & 3{,}717 & 3{,}428 & 1.084 \\
\bottomrule
\end{tabular}
\caption{\small Mixtral-8$\times$22B FP8, 4-node/32-GPU recompute/MBS sweep: all seven configurations projected and measured ($\rho=\tau=1.000$). Tag \texttt{rN\_bM} denotes $N$ recomputed layers and MBS $M$; all other settings are held fixed across the sweep. The offset $k$ stays within 1.08--1.18 (mean 1.128, cv 2.4\%), so no pair flips.}
\label{tab:mixtral-full}
\end{table}

\begin{table}[h]
\centering\small\setlength{\tabcolsep}{5pt}
\begin{tabular}{lccrcrrrr}
\toprule
\textbf{Tag} & \textbf{PP} & \textbf{VPP} & \textbf{Recomp.\ layers} & \textbf{MBS} & \textbf{GBS} & \textbf{Projected} & \textbf{Measured} & \textbf{$k$} \\
\midrule
qwen\_i39 (winner) & 4 & 3 & 1  & 2 & 2{,}048 & 4{,}418 & 4{,}380 & 1.009 \\
qwen\_i04          & 4 & 2 & 3  & 2 & 2{,}048 & 4{,}230 & 4{,}174 & 1.013 \\
qwen\_i06          & 4 & 2 & 6  & 2 & 2{,}048 & 3{,}839 & 4{,}156 & 0.924 \\
qwen\_i08          & 4 & 2 & 8  & 2 & 2{,}048 & 3{,}739 & 4{,}041 & 0.925 \\
qwen\_i21          & 4 & 2 & 23 & 2 & 2{,}048 & 3{,}481 & 3{,}984 & 0.874 \\
qwen\_i10          & 2 & 1 & 47 & 2 & 1{,}024 & 3{,}326 & 3{,}802 & 0.875 \\
qwen\_i27          & 4 & 2 & 23 & 1 & 1{,}024 & 2{,}277 & 3{,}312 & 0.688 \\
\bottomrule
\end{tabular}
\caption{\small Qwen3-235B-A22B FP8, 4-node/32-GPU: all seven configurations projected and measured ($\rho=\tau=1.000$). The offset $k$ drifts with configuration (1.01 at shallow recompute to 0.69 for the MBS=1, GBS=1{,}024 config) but stays small relative to the throughput gaps, so no pair flips.}
\label{tab:qwen-full}
\end{table}

\begin{table}[h]
\centering\small\setlength{\tabcolsep}{5pt}
\begin{tabular}{llrrr}
\toprule
\textbf{Tag} & \textbf{Distinguishing knobs} & \textbf{Projected} & \textbf{Measured} & \textbf{$k$} \\
\midrule
winner (trial32)     & mbs4, recompute none, sf1   & 13{,}206 & 12{,}168 & 1.085 \\
vpp6 (trial54)       & vpp6                        & 12{,}917 & 11{,}838 & 1.091 \\
pp4 (trial42)        & pp4                         & 12{,}693 & 11{,}593 & 1.095 \\
rec9 (trial29)       & mbs8, full recompute 9      & 12{,}403 & 11{,}029 & 1.125 \\
mbs16 (trial20)      & mbs16, full recompute 18    & 11{,}719 &  9{,}489 & 1.235 \\
baseline (reference) & mbs8, full recompute 18     & 11{,}260 & 10{,}109 & 1.114 \\
mbs2 (trial51)       & mbs2, recompute none        & 11{,}228 & 10{,}325 & 1.087 \\
mbs1 (trial23)       & mbs1, recompute none        &  9{,}247 &  7{,}598 & 1.217 \\
mbs1+rec18 (trial19) & mbs1, full recompute 18     &  6{,}781 &  6{,}509 & 1.042 \\
\bottomrule
\end{tabular}
\caption{\small GPT-OSS-120B BF16, 8-node/64-GPU: nine configurations, no prior AMD anchor. Top-1 projected $=$ top-1 measured ($+$20.4\% over the reference baseline, tracking $+$17.3\% projected).}
\label{tab:gptoss-full}
\end{table}

\begin{figure}[h]
\centering
\begin{tikzpicture}
\begin{groupplot}[
  group style={group size=2 by 1, horizontal sep=1.9cm},
  width=0.5\linewidth, height=4.8cm,
  tick label style={font=\scriptsize}, label style={font=\scriptsize},
  title style={font=\small}, scatter/use mapped color={draw=oiblue,fill=oiblue},
]
\nextgroupplot[
  title={Qwen3-235B ($\rho{=}\tau{=}1.000$)},
  xlabel={Projected tok/s/GPU}, ylabel={Measured tok/s/GPU},
  xmin=2000, xmax=4700, ymin=3000, ymax=4700,
  xtick={2500,3500,4500}, ytick={3000,3500,4000,4500},
]
\addplot[only marks, mark=*, mark size=2pt, oiblue] coordinates {
  (4418,4380) (4230,4174) (3839,4156) (3739,4041) (3481,3984) (3326,3802) (2277,3312)};
\addplot[dashed, gray, domain=3000:4600, forget plot] {x};
\node[font=\scriptsize, oiblue] at (axis cs:4418,4380) [above left] {winner};
\nextgroupplot[
  title={GPT-OSS-120B ($\rho{=}0.933$)},
  xlabel={Projected tok/s/GPU}, ylabel={Measured tok/s/GPU},
  xmin=6000, xmax=13800, ymin=6000, ymax=13000,
  xtick={7000,9000,11000,13000}, ytick={7000,9000,11000,13000},
]
\addplot[only marks, mark=*, mark size=2pt, oigreen] coordinates {
  (13206,12168) (12917,11838) (12693,11593) (12403,11029) (11719,9489)
  (11260,10109) (11228,10325) (9247,7598) (6781,6509)};
\addplot[dashed, gray, domain=6500:13000, forget plot] {x};
\node[font=\scriptsize, oigreen] at (axis cs:13206,12168) [above left] {winner};
\node[font=\scriptsize, oired] at (axis cs:11719,9489) [below right] {mbs16};
\end{groupplot}
\end{tikzpicture}
\caption{\small Projected vs.\ measured throughput per config (dashed
line $=$ identity). \textbf{Left:} Qwen3-235B is perfectly monotone
($\rho{=}\tau{=}1.000$)---the multiplicative offset $k$ shifts points
off the identity line but never reorders them. \textbf{Right:}
GPT-OSS-120B ($\rho{=}0.933$); the only reversals involve one
over-projected trio around \texttt{mbs16}, while the top-1 pick and the
extremes are exactly ordered (selection regret 0\%).}
\label{fig:rank}
\end{figure}
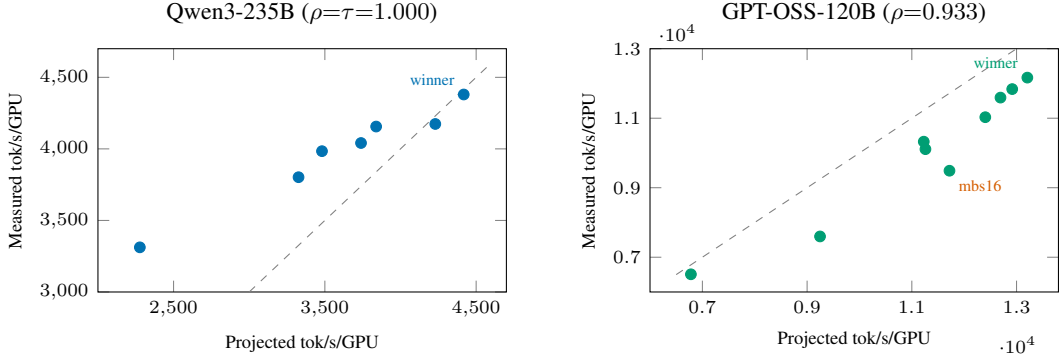

\paragraph{Why the GPT-OSS reversals occur, and why they are harmless.}
On $n=9$ there are 36 pairs; $\tau=0.833$ is 33 concordant and 3
discordant, all involving one trio---\texttt{mbs16}, the reference
\texttt{baseline}, and \texttt{mbs2}:
\begin{center}\small
projected: mbs16~11{,}719 $>$ baseline~11{,}260 $>$ mbs2~11{,}228\qquad
measured: mbs2~10{,}325 $>$ baseline~10{,}109 $>$ mbs16~9{,}489
\end{center}
The single driver is \texttt{mbs16} (trial20), over-projected at
$k=1.235$ vs.\ the cluster-typical 1.09--1.12: large-microbatch,
full-recompute configs have load-dependent activation-recompute overlap
that composing per-kernel benchmark timings does not fully capture,
inflating the projection and ranking it a few places too high. The
baseline-vs-\texttt{mbs2} swap is a genuine near-tie (0.2\% projected).
All three lie within $\sim$9\% of one another in measured throughput
(9{,}489--10{,}325); every pair separated by $>$$\sim$9\% is ordered
correctly, the top-4 and bottom-2 are exact, and selection regret is
0\%. Notably \texttt{mbs1} (trial23) has a comparably large offset
($k=1.217$) yet causes no reversal because it sits in a sparse part of
the range---a reversal requires \emph{both} a large offset and a
near-tie, which is why residual errors stay mid-pack.

\paragraph{Systematic ranking-consistency metrics.}
Table~\ref{tab:rank-metrics} reports, per workload, the discordant pairs
annotated with the measured gap each spans, the deviation of the
per-config offset $k$ from the workload median, a threshold-aware
concordance (pairs separated by $>$9\% measured throughput), and
selection (top-1) and top-3 regret. On GPT-OSS every reversal involves
the config whose $k$ deviates most from the median (\texttt{mbs16},
$+0.14$), and all reversals span measured gaps $\le$8.8\%, so concordance
is 100\% above that threshold. Mixtral's offsets all lie within 0.05 of
the median, so no pair is close to flipping. Qwen3's largest deviation
(\texttt{qwen\_i27}, $-0.24$) causes no reversal because it is isolated
at the bottom of the range.

\begin{table}[h]
\centering\small\setlength{\tabcolsep}{4pt}
\resizebox{\linewidth}{!}{%
\begin{tabular}{lcccccc}
\toprule
\textbf{Workload} & \textbf{Concordant} & \textbf{Discordant (meas.\ gap)} & \textbf{Median $k$} & \textbf{Max $|k-\tilde{k}|$} & \textbf{Conc.\ $>$9\%} & \textbf{Top-1/Top-3 regret} \\
\midrule
Mixtral-8$\times$22B & 21/21 & none & 1.132 & 0.048 (\texttt{r14\_b1}) & 100\% (10/10) & 0\% / 0\% \\
Qwen3-235B-A22B      & 21/21 & none & 0.924 & 0.236 (\texttt{i27}) & 100\% (10/10) & 0\% / 0\% \\
GPT-OSS-120B         & 33/36 & 2.1\%, 6.5\%, 8.8\% & 1.095 & 0.140 (\texttt{mbs16}) & 100\% (27/27) & 0\% / 0\% \\
\bottomrule
\end{tabular}}
\caption{\small Ranking-consistency metrics. Discordant pairs on GPT-OSS: baseline--\texttt{mbs2} (2.1\%), \texttt{mbs16}--baseline (6.5\%), \texttt{mbs16}--\texttt{mbs2} (8.8\%). Mixtral tags as in Table~\ref{tab:mixtral-full}.}
\label{tab:rank-metrics}
\end{table}

\clearpage
\section{MoE Load-Balancing Analysis}
\label{app:moe}

This appendix expands Section~\ref{ssec:moe-balance}. For benchmarking
we set \texttt{moe\_router\_force\_load\_balancing=true}: a random
straight-through router (forward assigns tokens uniformly at random;
backward still trains the real gate) plus an exactly even dispatcher
make executed load uniform (max/mean$=$1 by construction), so the
uniform-load assumption holds exactly for these runs. The flag is a
benchmarking device only and is never used in real training.

\paragraph{Three regimes, kept distinct.} Load balance is easy to
conflate across settings, so we separate them:
(i)~\emph{forced-balanced benchmark runs}, $I=1$ exactly;
(ii)~\emph{real training}, where aux-loss or aux-loss-free balancing plus
large-batch averaging keep $I$ near~1, since per-expert deviation shrinks
as $1/\sqrt{\text{tokens/expert}}$, giving a steady-state $I\approx1.1$--1.3;
and (iii)~\emph{deliberate stress tests} with balancing disabled, where we
push $I$ to 4$\times$, well beyond what training produces.

\paragraph{Stress test and correction.} With balancing off, two effects
appear: grouped-GEMM time is gated by the busiest EP-rank, i.e., by
$I_{\text{rank}}(E/\text{EP})$ rather than the mean, and the All-to-All
undergoes a regime jump. Sweeping $I$ from 1$\times$ to 4$\times$ on real
MI355X on three models (Table~\ref{tab:skew}), a uniform-load model
errs by up to 53--71\%, while a distribution-aware busiest-rank
correction keeps the error under 10\% (typically under 5\%) across the
full range. The $I\ge2$ points lie outside the training regime and are
included only to bound the worst case.

\begin{table}[h]
\centering
\begin{minipage}[t]{0.48\linewidth}
\centering\small\setlength{\tabcolsep}{5pt}
\captionof{table}{\small Skew sweep (1$\times$--4$\times$), FP8: worst-case error with and without the busiest-rank correction.}
\label{tab:skew}
\begin{tabular}{lrr}
\toprule
\textbf{Model} & \textbf{Uncorrected} & \textbf{Corrected} \\
\midrule
Qwen3-235B-A22B & 65.8\% & 3.3\% \\
Mixtral-8$\times$22B & 70.8\% & 6.1\% \\
DeepSeek~V3     & 53.1\% & 8.4\% \\
\bottomrule
\end{tabular}
\end{minipage}\hfill
\begin{minipage}[t]{0.48\linewidth}
\centering\small\setlength{\tabcolsep}{4pt}
\captionof{table}{\small Trained DeepSeek~V3 checkpoint (native learned balancing): per-layer vs.\ step-level pooled imbalance.}
\label{tab:trained-moe}
\begin{tabular}{lccr}
\toprule
\textbf{EP} & \textbf{experts/dev} & \textbf{per-layer (med)} & \textbf{pooled} \\
\midrule
8  & 32 & 1.38 & 1.05 \\
16 & 16 & 1.58 & 1.08 \\
32 & 8  & 2.13 & 1.12 \\
\bottomrule
\end{tabular}
\end{minipage}
\end{table}

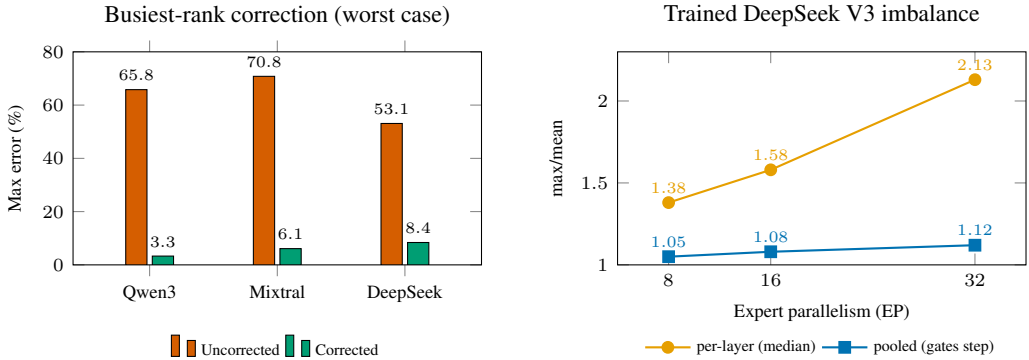
\begin{figure}[h]
\centering
\begin{tikzpicture}
\begin{groupplot}[
  group style={group size=2 by 1, horizontal sep=1.8cm},
  width=0.5\linewidth, height=4.4cm,
  tick label style={font=\scriptsize}, label style={font=\scriptsize},
  title style={font=\small},
]
\nextgroupplot[
  title={Busiest-rank correction (worst case)},
  ylabel={Max error (\%)}, ymin=0, ymax=80,
  ybar, bar width=8pt,
  symbolic x coords={Qwen3, Mixtral, DeepSeek},
  xtick=data, enlarge x limits=0.3,
  legend style={font=\scriptsize, at={(0.5,-0.30)}, anchor=north, legend columns=2, draw=none},
  nodes near coords, every node near coord/.append style={font=\scriptsize},
]
\addplot[fill=oired] coordinates {(Qwen3,65.8) (Mixtral,70.8) (DeepSeek,53.1)};
\addplot[fill=oigreen] coordinates {(Qwen3,3.3) (Mixtral,6.1) (DeepSeek,8.4)};
\legend{Uncorrected, Corrected}
\nextgroupplot[
  title={Trained DeepSeek~V3 imbalance},
  xlabel={Expert parallelism (EP)}, ylabel={max/mean},
  xtick={8,16,32}, xmin=4, xmax=36, ymin=1, ymax=2.3,
  legend style={font=\scriptsize, at={(0.5,-0.30)}, anchor=north, legend columns=2, draw=none},
  every axis plot/.append style={thick, mark size=2pt},
  nodes near coords, every node near coord/.append style={font=\scriptsize},
]
\addplot[oiorange, mark=*] coordinates {(8,1.38) (16,1.58) (32,2.13)};
\addplot[oiblue, mark=square*] coordinates {(8,1.05) (16,1.08) (32,1.12)};
\legend{per-layer (median), pooled (gates step)}
\end{groupplot}
\end{tikzpicture}
\caption{\small \textbf{Left:} disabling load balancing inflates
worst-case error to 53--71\%, but the busiest-rank correction brings it
back under 10\% on all three models. \textbf{Right:} on the trained
DeepSeek~V3 checkpoint the per-layer imbalance grows with EP, yet the
step-level \emph{pooled} imbalance---the quantity that actually gates
throughput---stays near~1.05--1.12 because the busiest device rotates
layer-to-layer.}
\label{fig:moe}
\end{figure}

The correction gates compute by the busiest EP-rank's load through the
measured grouped-GEMM scaling (a function of experts-per-rank
$E/\text{EP}$) and adds an A2A regime term when skew moves the largest
message across a bandwidth/latency boundary. On the trained checkpoint
(Table~\ref{tab:trained-moe}) the per-layer median at EP=8 is 1.38, but
the step-level \emph{pooled} imbalance is only 1.05 because the busiest
device rotates layer-to-layer: pooling each device's work over the step
before taking max/mean satisfies
$\max_r \sum_l n_{l,r} \le \sum_l \max_r n_{l,r}$, so per-layer peaks on
different devices partially cancel. In the pipelined/overlapped regime
step time tracks pooled per-device work, so the 1.05 value, not
1.38, governs throughput. Across EP=8/16/32 the pooled imbalance stays
at 1.05--1.12, within a few percent of uniform, and even the per-layer
median of 1.38 at EP=8 lies in the small-$I$ region of the sweep where
the corrected error is $\approx$2\%.
The three regimes order as the mechanism predicts (forced-balanced
1.00 $<$ trained 1.38 $<$ random-init 2.27), and because the penalty
depends only on $E/\text{EP}$ it transfers across EP (validated on
Mixtral, EP$\in\{2,4,8\}$, $\pm$6\%).

\section{Baselines: Calculon and Galvatron-Style Search}
\label{app:baselines}

\paragraph{Calculon (SC'23).} We first verified our Calculon~\cite{calculon}
setup against its own published NVIDIA A100 results (3.65\% avg /
8.87\% max error). We then ran the head-to-head on AMD hardware:
Llama-3.1-70B on MI325X (8~nodes, 64~GPUs, FSDP2, seq.\ length
8{,}192), comparing both tools' predictions with the measured
throughput (Table~\ref{tab:calculon}).
Primus is $\sim$4$\times$ more accurate; Calculon's opposite-direction
errors inflate the FP8--BF16 gap from 1.5$\times$ (measured) to
2.9$\times$. By source inspection, Calculon models no experts,
All-to-All, or ROCm software path, so Mixtral and DeepSeek~V3 cannot be
expressed at all.

\begin{table}[h]
\centering\small\setlength{\tabcolsep}{6pt}
\begin{tabular}{lrrr}
\toprule
\textbf{Config} & \textbf{Measured} & \textbf{Primus} & \textbf{Calculon} \\
\midrule
FP8  (GBS 256, MBS 4, FSDP) & 1{,}726 & 1{,}825 ($+$5.7\%)  & 2{,}006 ($+$16.2\%) \\
BF16 (GBS 64, MBS 1, FSDP)  & 1{,}174 & 1{,}086 ($-$7.5\%)  & 686 ($-$41.6\%) \\
\midrule
Mean $|$error$|$            & ---     & \textbf{6.6\%}      & 28.9\% \\
\bottomrule
\end{tabular}
\caption{\small Head-to-head accuracy on Llama-3.1-70B, AMD MI325X, 8~nodes / 64~GPUs (tok/s/GPU). Measured values are real MI325X runs.}
\label{tab:calculon}
\end{table}

\begin{figure}[h]
\centering
\begin{tikzpicture}
\begin{groupplot}[
  group style={group size=2 by 1, horizontal sep=1.5cm},
  tick label style={font=\scriptsize}, label style={font=\scriptsize},
  title style={font=\small},
]
\nextgroupplot[
  title={Ablation ranking (Mixtral, FP8)},
  width=0.52\linewidth, height=4.6cm,
  xbar, bar width=8pt, xmin=0, xmax=5600, clip=false,
  xlabel={Projected tok/s/GPU},
  symbolic y coords={{Galvatron-style b{=}30},{Tuning Agent+Sim},{Galvatron-style FP8},{Claude Code+Bench},{Tuning Agent+Bench}},
  ytick=data, y tick label style={font=\scriptsize},
  nodes near coords, every node near coord/.append style={font=\scriptsize},
  enlarge y limits=0.15,
]
\addplot[fill=oiblue] coordinates {
  (2155,{Galvatron-style b{=}30}) (3445,{Tuning Agent+Sim}) (3656,{Galvatron-style FP8})
  (4275,{Claude Code+Bench}) (4698,{Tuning Agent+Bench})};
\draw[dashed, oired, thick] (axis cs:3368,{Galvatron-style b{=}30}) -- (axis cs:3368,{Tuning Agent+Bench})
  node[pos=0, right, font=\scriptsize, oired] {untuned baseline 3{,}368};
\nextgroupplot[
  title={Prediction error vs.\ Calculon},
  width=0.4\linewidth, height=4.6cm,
  ybar, bar width=9pt, ymin=-50, ymax=25,
  ylabel={Signed error (\%)},
  symbolic x coords={FP8, BF16}, xtick=data, enlarge x limits=0.55,
  legend style={font=\scriptsize, at={(0.5,-0.28)}, anchor=north, legend columns=2, draw=none},
  nodes near coords, every node near coord/.append style={font=\scriptsize},
]
\addplot[fill=oigreen] coordinates {(FP8,5.7) (BF16,-7.5)};
\addplot[fill=oired] coordinates {(FP8,16.2) (BF16,-41.6)};
\draw[gray] (axis cs:FP8,0) -- (axis cs:BF16,0);
\legend{Primus, Calculon}
\end{groupplot}
\end{tikzpicture}
\caption{\small \textbf{Left:} on Mixtral-8$\times$22B, \agentname{}+Benchmark
(4{,}698) tops every simulate-oracle method---including the Galvatron-style
exhaustive 530-config FP8 search (3{,}656)---because the benchmark
oracle measures superposed kernel/collective effects. \textbf{Right:}
signed prediction error on Llama-3.1-70B; Calculon's large
opposite-sign errors ($+$16.2\% FP8, $-$41.6\% BF16) distort the
cross-precision trade-off, whereas Primus stays within $\pm$8\%.}
\label{fig:baselines}
\end{figure}
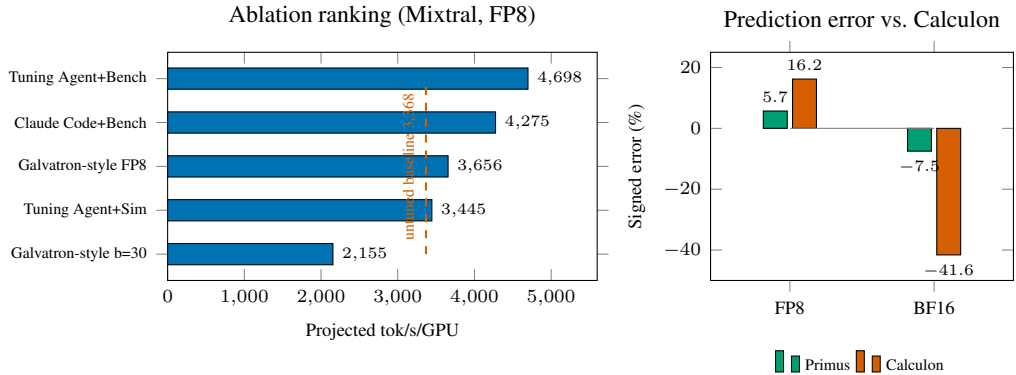

\paragraph{Galvatron-style DP search (PVLDB~16).} We implement a search
in the spirit of Galvatron~\cite{galvatron} over a 530-config space:
an outer loop over PP$\in\{1,2,4,7,8,14,28\}$, inner enumeration of all
(TP,EP,DP) triples with TP$\times$EP$\times$DP$=$world\_size$/$PP and
TP\,$\le$\,GPUs/node, crossed with legal MBS. Two adaptations make the
comparison apples-to-apples: (1) uniform parallelism across layers, as
Megatron-LM requires, reducing Galvatron's per-layer DP to full
enumeration; and (2) we grant the search our own Primus \emph{simulate}
oracle (analytical GEMM, tile-level FlashAttention-3, and an
algorithm-selecting per-collective model that also covers EP, MoE
All-to-All, and zero-bubble/ZBV/interleaved schedules that Galvatron
omits)---if anything an advantage to the baseline. Even so, the
exhaustive FP8 sweep (3{,}656) trails \agentname{}$+$Benchmark (4{,}698) by
22\%, i.e., the agent is 28\% higher (Table~\ref{tab:ablation}); the exhaustive BF16 sweep reaches
3{,}576. We ran these as two separate sweeps because precision is not a
knob in Galvatron's search space: it cannot move from BF16 to FP8
during a search the way our agent does. Started from the BF16 reference
with a 30-trial budget, as the agent is, it therefore stays in BF16 and
reaches only 2{,}155, below the untuned baseline (3{,}368). Fixing FP8
in advance removes that handicap, yet the gap remains. The gap is the
oracle, not the search: only real-hardware
benchmarking captures the superposed FP8 tensor-core, RCCL A2A, and
grouped-GEMM effects, and it ranks the winning EP=8 topology correctly
(Spearman $\rho=0.92$ over the $n=17$ configurations benchmarked on
hardware). Neither Galvatron nor Galvatron-BMW~\cite{galvatronbmw} evaluates MoE
architectures or reports projected-vs-measured rank correlation.

\begin{table}[h]
\centering\small\setlength{\tabcolsep}{4pt}
\resizebox{\linewidth}{!}{%
\begin{tabular}{lllrr}
\toprule
\textbf{Method} & \textbf{Oracle} & \textbf{Winning configuration} & \textbf{Proj.} & \textbf{vs.\ base} \\
\midrule
\agentname + Benchmark   & Benchmark & TP=1 PP=4 EP=8 MBS=2 FP8 DeepEP, no recompute & 4{,}698 & +39.5\% \\
Claude Code + Benchmark  & Benchmark & TP=1 PP=8 EP=4 MBS=2 FP8 DeepEP, no recompute & 4{,}275 & +26.9\% \\
Galvatron-style FP8 exhaustive  & Simulate  & TP=2 PP=2 EP=4 MBS=1 FP8, no recompute        & 3{,}656 & +8.6\% \\
Galvatron-style BF16 exhaustive & Simulate  & TP=2 PP=4 EP=4 MBS=1 BF16, no recompute       & 3{,}576 & +6.2\% \\
\agentname + Simulate    & Simulate  & TP=1 PP=4 EP=8 MBS=1 FP8, full recompute      & 3{,}445 & +2.3\% \\
Galvatron-style budget=30       & Simulate  & TP=8 PP=1 EP=4 MBS=2 BF16, no recompute       & 2{,}155 & $-$36.0\% \\
\bottomrule
\end{tabular}}
\caption{\small Winning configuration per method in the Mixtral-8$\times$22B ablation (4-node MI355X; baseline 3{,}368 projected tok/s/GPU; Table~\ref{tab:ablation}). Only the benchmark oracle identifies the intra-node EP=8 DeepEP topology with TP=1.}
\label{tab:ablation-configs}
\end{table}

\section{Tuning-Agent LLM Backbone and Prompts}
\label{app:prompts}

The agent uses two Anthropic Claude models via an OpenAI-compatible
LiteLLM gateway: Planner $=$ Claude~Sonnet~4.6
(\texttt{claude-sonnet-4-6}, ChainOfThought) and RLM $=$
Claude~Opus~4.6 (\texttt{claude-opus-4-6}, ReAct tool loop), with
\texttt{max\_tokens}=8000, \texttt{timeout}=300\,s, $\le$25
candidates/round, $\le$4 rounds. The verbatim instruction strings for
the two DSPy signatures follow; a per-cluster \texttt{extra\_guidance}
string carries model/cluster-specific hints and is logged per workload
so the exact conditioning text is auditable.

\begin{lstlisting}[caption={Planner (TuningPlanSignature) verbatim instruction.}]
Before touching any tools, design a structured search plan for tuning a
distributed-training configuration on a target cluster. You receive:
architecture (num_layers, hidden_size, MoE config, ...), cluster spec,
legal_axes, history_summary, extra_guidance, scratchpad. Produce a
concrete, numbered plan: which axes are most likely to matter for this
(model, cluster), which configurations should be tried in what order,
and what to do if they fail. Account for memory headroom, intra/inter-
node boundaries, and known trade-offs (e.g. MoE topk dominating
activations, EP-spans-nodes hurting A2A).

Outputs: rationale, hypotheses (<=5), candidate_configs (JSON, 5-8
configs over keys subset of {tp, pp, ep, cp, mbs, num_microbatches, gbs,
vpp, pp_schedule, enable_zero_bubble, recompute_granularity,
recompute_num_layers, overlap_grad_reduce, use_turbo_deepep,
sync_free_stage, target_ep_size, fp8, cross_entropy_loss_fusion,
use_distributed_optimizer, use_torch_fsdp2}), polish_plan. MoE priority
order: (1) DeepEP (~35% win), (2) SyncFree stage 2->3, (3) FP8 hybrid,
(4) recompute none/selective if memory fits, (5) ZB/ZBV schedules with
matching VPP, (6) coarse TP/PP/EP/CP reshapes, (7) num_microbatches >=
4*PP (or 2*PP for ZB/ZBV); gbs = mbs * num_microbatches * dp.
\end{lstlisting}

\begin{lstlisting}[caption={RLM (TuningSearchSignatureBenchmark) verbatim instruction.}]
Drive an iterative search for the best parallelization configuration
using real GPU benchmark projections only. Tools:
evaluate_benchmark(config)->metrics (primary; every new candidate goes
through it); evaluate_memory_only(config); get_history(k=30);
get_best(); get_legal_axes(); get_architecture(); get_cluster();
get_budget_status(); note_to_scratchpad(text); read_scratchpad();
query_llm(prompt). Workflow each iteration: (1) read_scratchpad +
get_history to ground yourself; (2) pick the next candidate from the
plan not already in history; (3) evaluate_benchmark directly (if
legal=False, read reason and adjust); (4) after 3-5 evals, note what you
learned; (5) STOP when the benchmark budget is exhausted or the
incumbent has not improved in the last 5 trials.

Outputs: best_config (JSON), summary (5-8 sentences), next_steps.
\end{lstlisting}

\section{Limitations}
\label{app:limitations}

\paragraph{Network contention is not modeled.}
The communication model assumes dedicated bandwidth per topology domain.
Concurrent collectives competing for the same xGMI or NIC bandwidth
(e.g., TP AllReduce overlapping with FSDP AllGather on the same node)
will see effective bandwidth lower than the model assumes. We have not
yet observed this as a dominant error source on the configurations
evaluated here, but it could add error in some settings at extreme scale.

\paragraph{Reduced-parallelism fidelity has limits.}
Sub-node benchmarking necessarily measures a system whose parallelism
configuration differs from the target. The +6.6\% Mixtral error at the
2-GPU benchmark and the corresponding compounding pattern in
Section~\ref{sec:eval} bound this in practice; longer reductions (e.g.,
benchmarking PP=4 EP=8 from PP=1 EP=1) introduce more analytical work
and correspondingly more residual error. The framework reports both the
measured base time and the analytical delta for each restored dimension
so that users can audit which step contributed to the residual.

\paragraph{Hardware homogeneity assumption:}
All nodes are assumed identical, both in static specification (HBM, CU
count, link topology) and in runtime behavior (clock frequency under load).
GPU-to-GPU frequency variation within a job and partial-rack failure
modes are outside the model's scope.

\paragraph{\agentname search space:}
The deterministic seed planner enumerates a fixed set of axes and the
LLM phase explores combinations of those axes; the agent does not
introduce optimizations that are not exposed as flags by the projection
oracle. Adding a new optimization (e.g., a new MoE A2A backend) requires
extending both the oracle and the planner. The search space exposes
ZeRO-1 (\texttt{use\_distributed\_optimizer}) and ZeRO-3
(\texttt{use\_torch\_fsdp2}); ZeRO-2 is excluded because wherever it
would beat ZeRO-1, FSDP2 is available.

\paragraph{Precision is a throughput knob only.}
The agent treats precision (BF16/FP8) as a search axis for throughput
evaluation. Projection says nothing about numerical quality: FP8 recipes
must be validated for convergence before production use.

\paragraph{New layer types and accelerators.}
Extending beyond the evaluated layer types requires only a
parameterizable layer: subquadratic layers (e.g., Gated DeltaNet) use a
profiled FLOP estimate, and sparse attention a density-weighted FLOP
count; validating these is future work. Porting to NVIDIA requires a
hardware YAML and the 8-GPU benchmark; TPUs additionally require
adapting the collective model to XLA/JAX collectives.

\section{Broader Impact}
\label{app:impact}

\sysname is an infrastructure tool aimed at reducing the GPU-hours
consumed by configuration tuning for large model training. The intended
positive impact is twofold:
\begin{itemize}
\item \textbf{Energy and cost savings.} By converting much of the
  configuration search from multi-node trial-and-error into single-node
  projection, \sysname reduces the GPU-hours required to find a high-
  throughput configuration. The Mixtral case study in
  Section~\ref{ssec:agent-eval} consumed $\sim$30~minutes of single-node
  agent time plus a small number of validation runs, in place of what
  would otherwise be days of manual multi-node sweeps.
\item \textbf{Lower barrier to large-scale training.} Smaller research
  groups without standing access to large clusters can use the
  simulation backend to explore parallelism trade-offs and the agent to
  generate candidate configurations, then validate only a few selected
  candidates on rented capacity.
\end{itemize}
We are not aware of foreseeable negative societal impacts specific to
this tooling beyond those of distributed LLM training in general.
The framework is released under an open-source license to ensure that
the projection oracle, the seed planner, and the LLM agent loop are
auditable and modifiable by the community.

\end{document}